\documentclass{article}

\usepackage[preprint]{neurips_2024}

\PassOptionsToPackage{options}{natbib}

\usepackage[utf8]{inputenc} %
\usepackage[T1]{fontenc}    %
\usepackage{hyperref}       %
\usepackage{url}            %
\usepackage{booktabs}       %
\usepackage{amsfonts}       %
\usepackage{nicefrac}       %
\usepackage{microtype}      %
\usepackage{xcolor}         %
\usepackage{tcolorbox}

\usepackage{amsmath,amsfonts,bm}

\def\eqref#1{equation~\ref{#1}}

\def\1{\bm{1}}

\def\vw{{\bm{w}}}
\def\vx{{\bm{x}}}
\def\vy{{\bm{y}}}
\def\vz{{\bm{z}}}

\def\mI{{\bm{I}}}

\DeclareMathAlphabet{\mathsfit}{\encodingdefault}{\sfdefault}{m}{sl}
\SetMathAlphabet{\mathsfit}{bold}{\encodingdefault}{\sfdefault}{bx}{n}

\usepackage{paralist}

\usepackage{mathtools}
\usepackage{subcaption}
\usepackage{comment}
\usepackage{amsthm}
\newtheorem*{remark}{Remark}
\def\tv{{\bm{\tau}}}
\def\Nd{{N}}  %

\title{Task Vectors in In-Context Learning: \\ Emergence, Formation, and Benefits}

\author{Liu Yang$^w$, Ziqian Lin$^w$, Kangwook Lee$^w$, Dimitris Papailiopoulos$^{w,m}$ \& Robert D. Nowak$^w$ \\
$^w$University of Wisconsin-Madison, $^{m}$Microsoft Research \\
\texttt{\{liu.yang, zlin284, kangwook.lee\}@wisc.edu,} \\
\texttt{dimitris@papail.io, rdnowak@wisc.edu} \\
}

\usepackage{tcolorbox}

\newtcolorbox{highlight}[1][]{
    colback=yellow!10,
    colframe=gray!30,
    boxrule=0.1pt,
    arc=2pt,
    leftrule=1pt,
    rightrule=1pt,
    toprule=1pt,
    bottomrule=1pt,
    #1
}

\begin{document}

\tcbset{colback=lightgray!20, colframe=black, boxrule=0.5mm, arc=4mm, width=\linewidth}

\maketitle

\addtocontents{toc}{\protect\setcounter{tocdepth}{0}}
\begin{abstract}

In-context learning is a remarkable capability of transformers, referring to their ability to adapt to specific tasks based on a short history or context.  
Previous research has found that task-specific information is locally encoded within models, though their emergence and functionality remain unclear due to opaque pre-training processes.
In this work, we investigate the formation of task vectors in a controlled setting, using models trained from scratch on synthetic datasets. Our findings confirm that task vectors naturally emerge under certain conditions, but the tasks may be relatively weakly and/or non-locally encoded within the model.
To promote strong task vectors encoded at a prescribed location within the model, we propose an auxiliary training mechanism based on a \emph{task vector prompting loss (TVP-loss)}. 
This method eliminates the need to search for task-correlated encodings within the trained model and demonstrably improves robustness and generalization.
\end{abstract}

\section{Introduction}

To understand the underlying mechanisms of in-context learning in transformers, researchers have probed pre-trained models from various perspectives, such as altering the labels in demonstrations~\citep{Min2022RethinkingTR,Kim2022GroundTruthLM} and investigating circuit mechanisms~\citep{elhage2021mathematical,Wang2023LabelWA,Hanna2024HaveFI,Singh2024WhatNT}. Additionally, controlled, small-scale studies have been conducted by training transformers from scratch to observe their in-context learning behavior on linear regression tasks~\citep{Garg2022WhatCT,Oswald2022TransformersLI,lin2024dual}, discrete functions~\citep{Bhattamishra2023UnderstandingIL}, hidden Markov chains~\citep{Xie2021AnEO}, and DFAs~\citep{Akyrek2024InContextLL}. Furthermore, theoretical approaches have also been applied to this problem~\citep{Xie2021AnEO,lin2024dual,Giannou2023LoopedTA}.

Among the various efforts to probe pre-trained models, one significant line of research employs the concept of a {``task vector''}, which is a vector in the model's activation or weight space that encodes task-specific information. 
The concept of the {task vector} was first introduced by \cite{Ilharco2022EditingMW}, where it is defined as a direction in a model's weight space corresponding to a particular task. 
Subsequently, \cite{Hendel2023InContextLC} demonstrated that, given a task demonstration as context, a pre-trained large language model forms a task vector in its activation space\footnote{The ``activation space'' refers to the space where the output of each transformer layer resides.} at certain layers. 
This task vector encodes only the task information and is independent of the specific demonstration of the task. %
By inserting the task vector directly into the model, it is able to perform the task without context or demonstration (i.e., zero-shot). We will refer to this as \emph{Task Vector Prompting} (TVP) in this paper.
Concurrently, \cite{Liu2023IncontextVM, Todd2023FunctionVI, Merullo2023LanguageMI, Li2024ImplicitIL,saglam2024learning} have also identified a single vector that encodes the task information, albeit using different terminology.
We omit the details here and refer the reader to the related work and the original papers for more information.

\begin{figure}[t]
    \centering
    \includegraphics[width=0.9\linewidth]{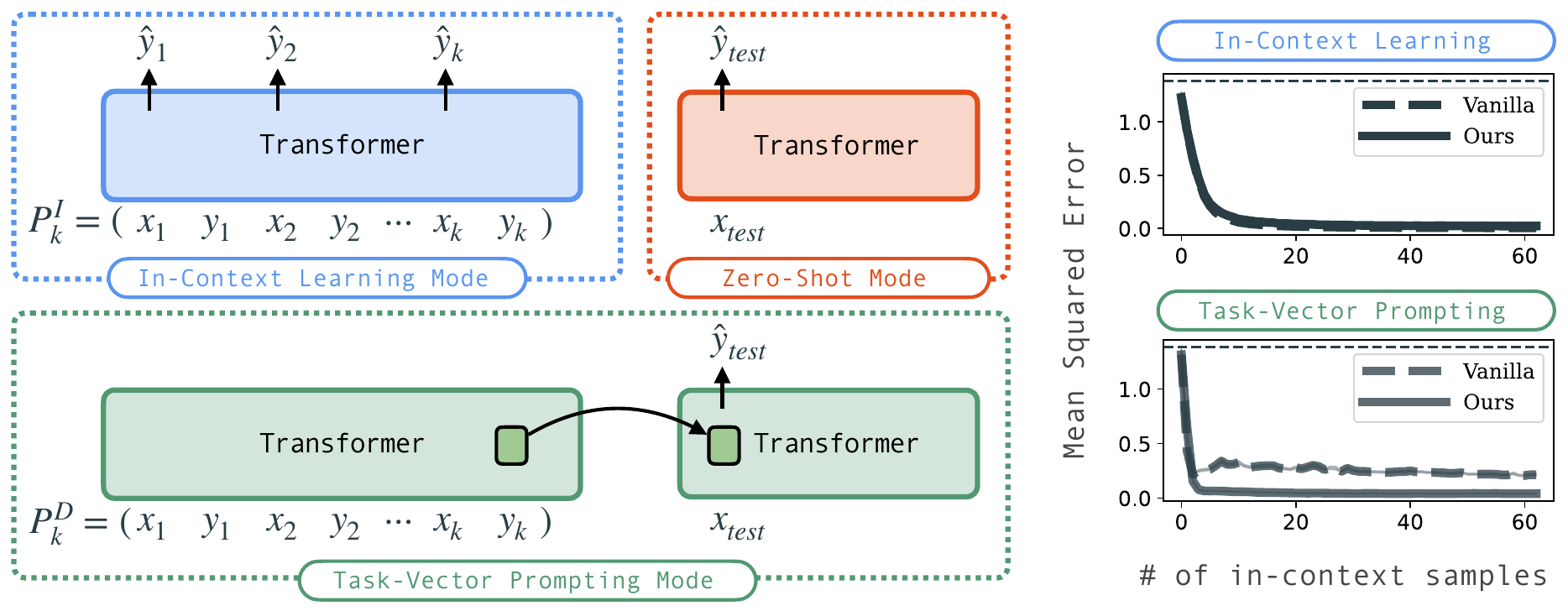}
    \caption{\small \emph{Overview of the transformer operating in in-context learning (ICL) and task vector prompting (TVP) modes.} A transformer can be configured to operate in ICL mode, using input-output pairs as prompts, or in TVP mode, extracting task-specific embeddings for zero-shot predictions (Architecture and training details in Section~\ref{sec:def}). On the right, the ICL and TVP performances of the vanilla-trained model and our method are shown, with the dashed horizontal line indicating random prediction performance (i.e., no task information is inferred). Compared to vanilla training, our approach enhances task-specific representations in the TVP mode while preserving comparable ICL performance.}
    \label{fig:overview_icl_tv}
\end{figure}

To better understand the emergence of task vectors, we examine in-context learning behavior in a controlled setting, training models from scratch on various synthetic datasets. 
As shown in Figure~\ref{fig:overview_icl_tv}, when trained from scratch on the linear regression task defined as $y_i = \vw^T\vx_i$, the transformer (dashed line) demonstrates the ability to use in-context learning (ICL) to solve the task (top-right).
We evaluate the trained model’s performance in task vector prompting (TVP) mode (Figure~\ref{fig:overview_icl_tv}, bottom-left), where the task vector is extracted from the in-context learning mode and injected back in a zero-shot manner, as defined by~\citet{Hendel2023InContextLC}.  The TVP performance is much better than chance, but a bit worse than ICL performance, which we attribute to the fact that the encoding of the task may not be strong and localized by normal training methods.
To encourage the formation of a strong and localized task vector, we propose an auxiliary training loss, called the \emph{task vector prompting loss (TVP-loss)}. In our new approach, the model is trained using the TVP-loss in addition to normal training losses.
As illustrated in Figure~\ref{fig:overview_icl_tv}, our approach (solid line) achieves comparable ICL performance to the vanilla model in the ICL mode (top-right) and its TVP performance is significantly improved and comparable to ICL performance with multi-shot demonstration.

The remainder of the paper is organized as follows. Section~\ref{sec:def} provides the formal definition of task-vector prompting. In Section~\ref{sec:task_vector_in_vanilla}, we analyze the characteristics of task vectors in trained-from-scratch transformers and identify that the encoded task vectors remain noisy. Section~\ref{sec:training_algo} introduces a training algorithm to encourage the formation of task vectors. Finally, we extend our analysis to synthetic formal language tasks in Section~\ref{sec:task_vector_formal_language}.
Our main contributions and findings are outlined below:

\paragraph{Emergence of Task Vectors During Training.} 
We investigate the effectiveness of task vector extraction methods, originally proposed for pre-trained large language models, when applied to small-scale models trained from scratch on synthetic datasets. Our study reveals that task vectors can naturally emerge during training, provided that specific input formats are used and the model has sufficient capacity. To understand the conditions that promote this emergence, we examine different hyper-parameters such as the model depth, the use of sparse attention, and input embedding sizes.

\paragraph{Strong Task Vectors with Proposed Auxiliary Loss.}
Although task vectors naturally emerge, they are often weak and entangled with information from input queries. This prevents task vectors from representing purely task-specific knowledge, as noted by~\citet{Hendel2023InContextLC}. To overcome this limitation, we propose a training algorithm that explicitly encourages the formation of task vectors independent of query information. This approach strengthens the task vector, ensuring that task-specific encoding is explicitly established within the model.  Another benefit is that our approach specifies the embedding location of the task vector within the model, eliminating uncertainty about where it might emerge.

\paragraph{Task Vectors for In-Context Learning Robustness.}
Using our proposed training algorithm, we analyze models trained on synthetic tasks, both with and without the enhanced task vectors at certain layers. We then assess the advantages of having strong task vectors in the model, demonstrating their robustness in in-context learning performance and examining the effects of task vectors forming in different layers.

\section{Task Vector Definition}
\label{sec:def}
In this section, we formally define the notion of task vector. %
Let $\mathcal{F}$ denote a class of functions or ``tasks", and let $\mathcal{X}$ and $\mathcal{Y}$ be the input and output spaces, respectively. 
If $f \in \mathcal{F}$ is the task in a specific ``context", then for any input $\vx \in \mathcal{X}$, the corresponding output is $\vy = f(\vx) \in \mathcal{Y}$.
Consider $\vx_{\text{test}} \in \mathcal{X}$, and the transformer model $M$.
We can measure the model's zero-shot performance by $\ell(f(\vx_{\text{test}}), M(P_{\text{query}}))$, where $P_{\text{query}} = [\vx_{\text{test}}]$ denotes the query input. The in-context learning performance with $k$-shot examples is measured by $\ell(f(\vx_{\text{test}}), M(P_k))$, where $P_k$ is the in-context prompt. The prompt contains the $k$-shot examples $[\vx_1, f(\vx_1), \cdots, \vx_k, f(\vx_k), \vx_{\text{test}}]$.

Now consider a \emph{demonstrated prompt} $P_k^D$, where the superscript $D$ highlights its role in demonstrating task information. This prompt contains $k$ in-context samples:
\[
P_k^D = [\vx_1, f(\vx_1), \cdots, \vx_k, f(\vx_k)] %
\]
A task vector is an internal embedding $\tv$ extracted when the model is presented with $P_k^D$, which encodes the task at hand ($f$). Inserting the task vector into the model during zero-shot prompting is denoted by $M(P_{\text{query}};\tv)$. 
It is worth noting that the task vector $\tv$ is extracted from the internal embeddings when inputting $P_k^D$, not $P_k$, therefore the task vector $\tv$ does not have explicit knowledge of $\vx_{\text{test}}$ when encoding the task.

A task vector extractor $g$ tries to locate this single embedding $\tv$ in the model given the demonstrated prompt $P_k^D$, i.e. $\tv = g(M(P_k^D))$. Then the performance of this extractor can be measured by $\ell(f(\vx_{\text{test}}), M(P_{\text{query}};\tv))$. A task vector is considered to be successfully formed in the model $M$ for the task $f$ if the performance of the extracted task vector, i.e. task vector prompting performance, closely aligns with the in-context learning performance and surpasses the zero-shot performance.

\citet{Hendel2023InContextLC} confirmed that, in pre-trained large language models, $g(M(P_k^D))$ corresponds to the output embedding at approximately the middle layers of the model during the forward pass, which is responsible for predicting $f(\vx_{\text{test}})$. Similarly, \citet{Sia2024WhereDI} found that the task in context is recognized by the model during the middle stage of the forward pass.

\section{Emergence of Task Vectors in Trained-from-Scratch Models}\label{sec:task_vector_in_vanilla}
In this section, we investigate whether task vectors can be identified in a transformer model trained from scratch on synthetic tasks.

\subsection{Experimental Setup}\label{sec:exp_lr}
We investigate the localization of the task vector using the GPT-2 decoder model~\citep{Radford2019LanguageMA} trained on the generated random prompts in the following tasks: 
\paragraph{\emph{Linear Regression.}} 
Following the setup in~\citep{lin2024dual,Garg2022WhatCT}, the input vectors $\vx_i \in \mathbb{R}^d$ are sampled from a normal distribution $\mathcal{N}(0, \mI)$. The linear function weights $\vw \in \mathbb{R}^d$, which define the task, are sampled from a mixture of $d$ Gaussians, each with means corresponding to the standard basis vectors: $\boldsymbol{\mu}_i \in \mathbb{R}^d$ with $\boldsymbol{\mu}_{ij} = 1$ if $i = j$, and $0$ otherwise. All Gaussians share a covariance matrix $\Sigma = \frac{1}{4} \mI$, and each component is selected with equal probability. The target outputs are computed as $f(\vx_i) = \vw^\top \vx_i$. We set $d=6$ in our experiments. 
\paragraph{\emph{Sinusoidal Regression.}} Following the same sampling strategy as the linear regression task, the target outputs are defined as $f(\vx_i) = \sin(\vw^\top \vx_i)$. In this task, \(\vw\) determines the underlying function.
\paragraph{\emph{Discrete Token Offset Prediction.}} In this task, we set the vocabulary size to be $C$, and each input token $\vx_i$ is a discrete value drawn from the set $\{0, 1, \dots, C-1\}$. The target output is defined as $f(\vx_i) = (a \times \vx_i + b) \bmod C$, where $a \in \{1, 2, 3\}$, $b \in \{0, 1, 2\}$ are uniformly sampled, and $\bmod$ denotes the modulo (remainder) operation. The task is uniquely determined by the choice of $a$ and $b$, resulting in total 9 possible tasks. By default, we set $C = 1000$.

During training, prompts are generated on the fly. Specifically, for each prompt, a function $f \in \mathcal{F}$ is randomly sampled according to the function distribution described for each task above. Subsequently, input tokens \(\{\vx_i\}_{i=1}^k\) are independently sampled from the corresponding input distribution. The function $f$ is then evaluated on these inputs to produce the target outputs, forming the in-context learning prompt \(P_k = [\vx_1, f(\vx_1), \dots, \vx_k, f(\vx_k), \vx_{\text{test}}]\). Let the distribution of such prompts be denoted as $\mathcal{P}$.
The transformer $M$ parameterized by $\theta$ is then trained to minimize the following expected loss:
$$
\min_{\theta}\mathbb{E}_{P_k \sim \mathcal{P}} \  \sum_{i=1}^{k} \ell(M_{\theta}(P_i), f(\vx_{i+1})).
$$
where $P_i = [\vx_1, f(\vx_1), \dots, \vx_i, f(\vx_i), \vx_{i+1}]$ represents the prompt prefix containing $i$ in-context examples.
In our experiments, we use GPT-2 model with an embedding size of 64, 4 attention heads, and 3 layers, trained with a maximum context length of 63 (i.e., $k=63$) and without positional embeddings (NoPE).
We use the Adam optimizer with a learning rate of 0.0001 and a batch size of 256, training for 300k iterations. With dynamically generated prompts, this corresponds to 76.8 million distinct prompts. All experiments are run on an NVIDIA GeForce RTX 3090.

We investigate various input formats and task vector extraction methods (details in Appendix~\ref{appendix:prompt_format} and Appendix~\ref{appendix:tv_extractor}) to identify conditions under which task encoding emerges in trained-from-scratch transformers. Based on this investigation, we focus on the following setup, where task vectors emerge most distinctly.:
\begin{itemize}
\itemindent=-20pt
    \item \emph{input format}: Prompts with $k$ in-context examples are generated in the format $P_k = [\vz, \vx_1, f(\vx_1), \dots, \vx_k, f(\vx_k), \vx_{\text{test}}]$, where $\vz$ is a special token placed at the beginning of the prompt to serve as a placeholder for injecting task encoding during zero-shot task vector prompting. During training, the embedding of $\vz$ is treated as a learnable parameter that is shared across all prompts.
    \item \emph{task vector extractor method}: inspired by~\citet{Hendel2023InContextLC}, we locate the task vectors \emph{at the activations of the token $\vz$ and $\{y_i\}$} in the input format mentioned above. Specifically, during task vector prompting,  the embedding is copied into the corresponding position in a zero-shot model, and the zero-shot loss is measured. A lower zero-shot loss indicates that the embedding effectively encodes task-specific information
    (details in Appendix~\ref{appendix:prompt_format}). 
\end{itemize}

As noted in~\citet{Hendel2023InContextLC}, the extraction process involves identifying the layer that optimally encodes the task vector. 
The optimal layer index $l^\star$ is determined by evaluating task vectors extracted from each layer and selecting the one that minimizes the average loss $\ell$ across all $\Nd$ demonstrated prompts. This procedure ensures that the selected layer provides the most effective task vector for predicting the test outputs.

Formally, for each task $f$, we generate $\Nd$ demonstrated prompts $\{P_{k}^{D,j}\}_{j=1}^{\Nd}$, where the same function $f$ is consistently used across all \(\Nd\) prompts, and the input to each prompt $P_{k}^{D,j}$ is uniformly sampled.
Let \(P_{\text{query}}^{j}\), with test input \(\vx_{\text{test}}^{j}\), represents the corresponding test queries for the $j$-th demonstrated prompt. 
For a task vector extracted from a $k$-shot prompt, the optimal layer index $l^\star$ is computed as:
$$
l^{\star} = \arg\min_l \sum_{j=1}^{\Nd}\ell\left(f(\vx_{\text{test}}^{j}), M\left(P_{\text{query}}^{j};g(M(P_{k}^{D,j}), l)\right)\ \right)
$$
where $g(M(P_{k}^{D,j}), l)$ denotes the task vector extracted from the $l$-th layer when processing the demonstrated prompt $P_{k}^{D,j}$, and $M(P_{\text{query}}^{j}; \tv)$ represents the model’s prediction for the query prompt $P_{\text{query}}^j$, where the task vector $\tv$ extracted from the $l$-th layer is copied to the corresponding position in the model’s representation when processing $P_{\text{query}}^j$. A lower loss $\ell$ indicates that the corresponding layer provides a more effective encoding of task-specific information.

\begin{remark}[Task Vector Location]
    The work of~\citet{Hendel2023InContextLC} employs an in-context learning prompt format structured as $P_k = [\vx_1, \vz, f(\vx_1), \cdots, \vx_k, \vz, f(\vx_k), \vx_{k+1}, \vz]$, where $\vz$ indicates the ``maps-to'' token. This setup differs from the prompt format used in our experiments. As detailed in Appendix~\ref{appendix:prompt_format}, we investigate various input formats to determine their effect on task vector emergence. We find that the trained-from-scratch model exhibits task vector emergence only when the input format for training is $P_k = [\vz, \vx_1, f(\vx_1), \cdots, \vx_k, f(\vx_k), \vx_{k+1}]$, where no additional tokens are placed between $\vx$ and $f(\vx)$. 
    
    We hypothesize that the discrepancy arises because pre-trained LLMs leverage the semantic meaning of the ``maps-to'' token as an anchor for task summarization, whereas in trained-from-scratch transformers, this extra ``maps-to'' token serves only as a computational placeholder, with fixed token positions negating the need for a delimiter.
\end{remark}

\begin{figure}[ht]
    \centering
    \includegraphics[width=0.9\linewidth]{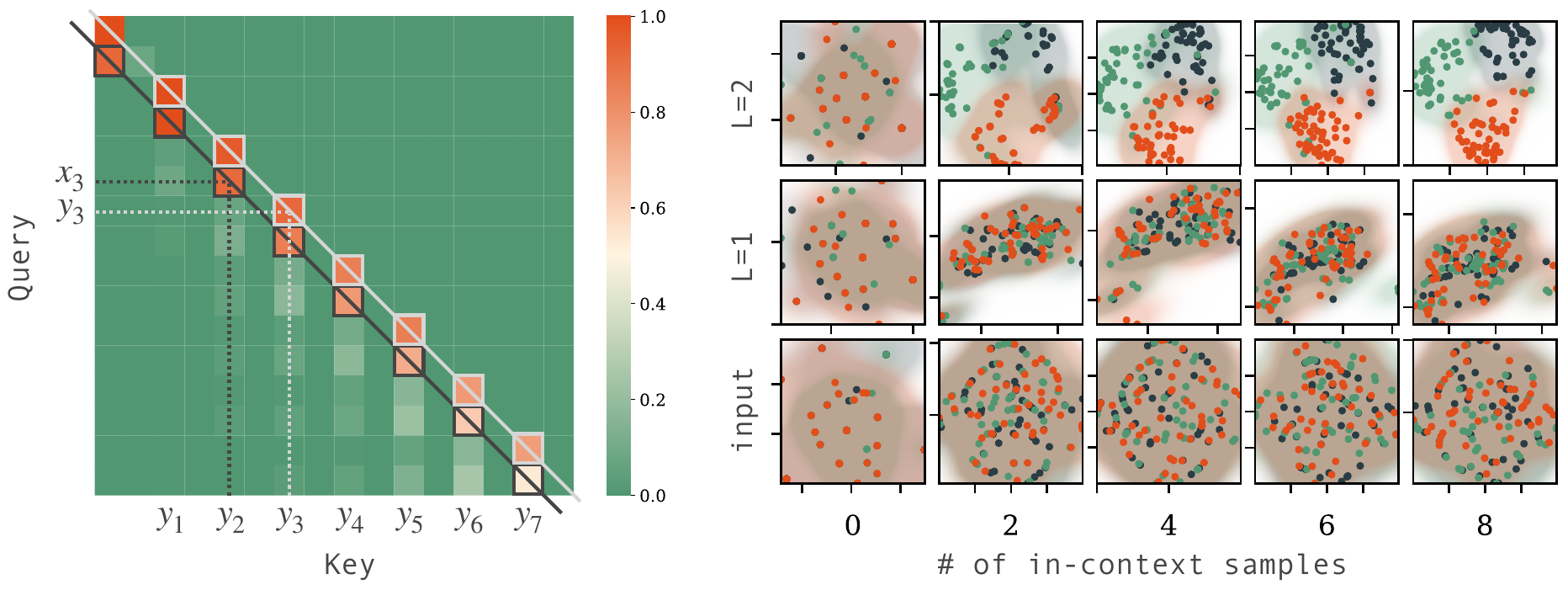}
    \caption{\small \emph{Attention map and PCA visualization of activations across three different linear functions.} 
    (\textbf{Left}) For the linear regression task described in Section~\ref{sec:exp_lr}, the attention map illustrates how each query (rows) attends to all available keys (columns), with each row summing to 1.  The heatmap reveals that the activations at the $\vx_i$ positions predominantly attend to the activations of the preceding $y_{i-1}$, where task information is stored (highlighted by the black boxes). Additionally, the activations at the $y_i$ positions attend to both themselves and the preceding $y_{i-1}$, enabling the online updating of task information (highlighted by the white boxes).
    (\textbf{Right}) PCA visualizations of token $y_i$ activations ($i \in \{0, 2, 4, 6, 8\}$) across layers $L$ reveal that task-specific clusters (three colors correspond to three different tasks) begin to emerge at the output of the 2nd layer, indicating that the model progressively encodes task information as depth increases. }
    \label{fig:vanilla_cluster}
\end{figure}

\subsection{Task Vector Emergence}

\begin{highlight}
    \paragraph{Finding 1:} 
    \emph{Task vectors naturally emerge in small models trained from scratch.}
\end{highlight}

In Figure~\ref{fig:overview_icl_tv}, we demonstrate the presence of task encoding in the activations of token $y_i$ under the linear regression task, using the experimental setup described in Section~\ref{sec:exp_lr} with number of demonstrated prompts $\Nd=50$. 
In the task vector prompting mode, the transformer achieves a mean squared error (MSE) of approximately 0.25, significantly lower than the random prediction baseline of around 1.2, indicating that the model successfully infers task information in its representation. 
Throughout the paper, we use \(\Nd=50\) by default unless stated otherwise. Additionally, we examine the impact of different \(\Nd\) values on task vector prompting performance, as detailed in Appendix~\ref{appendix:vanilla_varying_N}.

To understand the non-random performance observed in the task vector prompting mode (Figure~\ref{fig:overview_icl_tv}), we analyze how the model encodes task information using attention maps (\emph{left}) and a PCA of activations at each layer while varying the number of in-context examples (\emph{right}), as shown in Figure~\ref{fig:vanilla_cluster}. In transformer models, attention mechanisms enable each input token to dynamically focus on specific parts of the input data by assigning importance to them. This process can be visualized using attention maps, which quantify the relationships between input tokens. These maps are computed using query and key vectors: the query represents the current focus of interest, while the key interact with the query to calculate importance scores (via their dot product), determining how relevant each input element is to the query.

Recall that we follow the experimental setup in Section~\ref{sec:exp_lr} and conduct experiments on the linear regression task using a 3-layer transformer with a 63-shot context. On the left, the attention map of a specific head in 2nd layer shows that the activations at $\vx_i$ position primarily attend to the activations of the preceding $y_{i-1}$, where task information is stored. Furthermore, the activations at the $y_i$ location attend to both themselves and the preceding $y_{i-1}$, enabling the model to update task information online. On the right, we present a PCA of the activations for token $y_i$ ($i \in \{0, 2, 4, 6, 8\}$) across layers $L$, evaluated on three different linear functions: $\vw_1 = [1, 0, 0, 0, 0, 0]^T$, $\vw_2 = [0, 0, 1, 0, 0, 0]^T$, and $\vw_3 = [0, 0, 0, 0, 1, 0]^T$. 
At the input to the third layer, the activations for different tasks begin to form distinct clusters, though there is still considerable overlap and dispersion among the clusters. %
We also perform this analysis for the input format used in~\citet{Hendel2023InContextLC} and present the results in Figure~\ref{fig:x->y_attn_cluster} in Appendix~\ref{appendix:prompt_format}. In this format, we found no evidence of task encoding.

In the following sections, we analyze the impact of model depth and context length on task vector emergence across different problem dimensions (Section~\ref{sec:vanilla_vary_d}) and investigate the locations of task vectors (Section~\ref{sec:vanilla_tv_layer}). Additionally, we explore the effects of embedding size and sparse attention in Appendices~\ref{appendix:model_capacity} and~\ref{appendix:sparse_attention}, respectively.

\subsubsection{Effects of Model Depth and Context Length on Task Vector Emergence}\label{sec:vanilla_vary_d}

\begin{figure}[ht]
    \centering
    \includegraphics[width=0.85\linewidth]{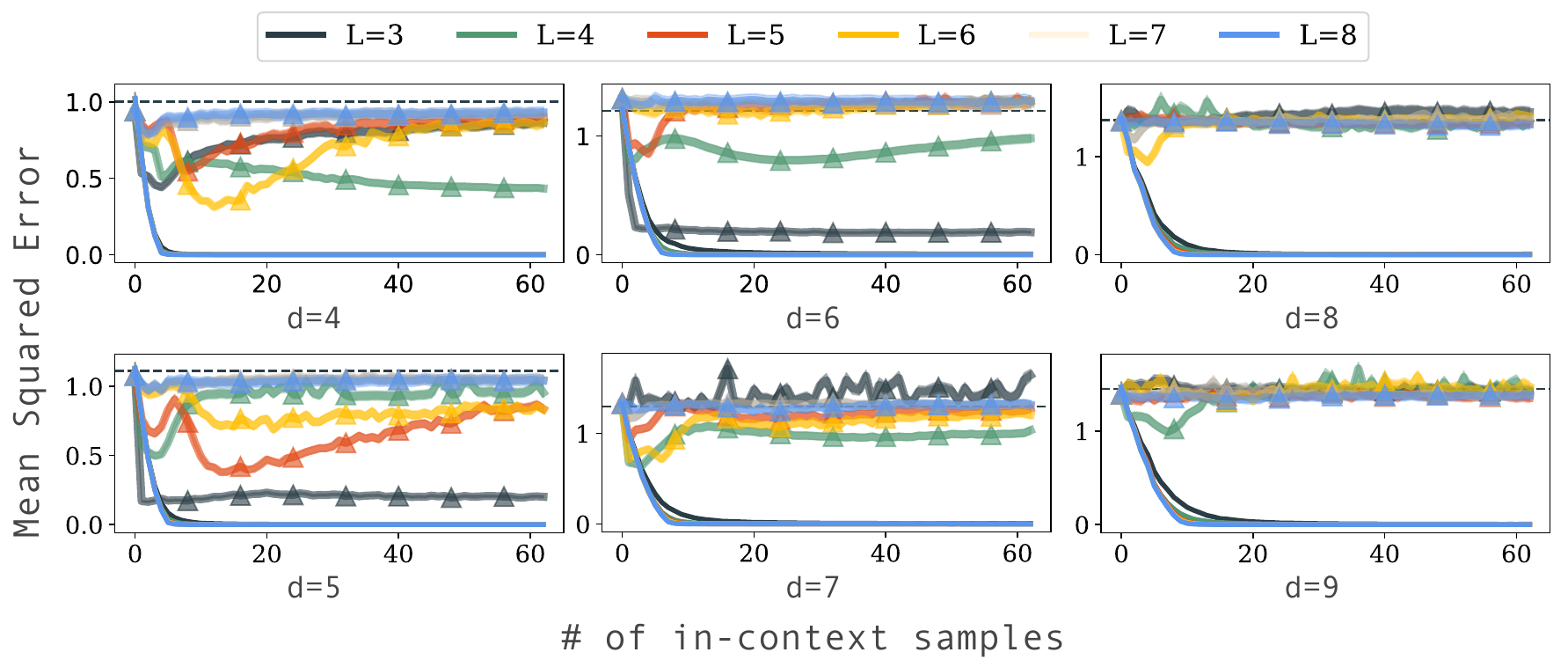}
    \caption{\small \emph{Performance of the transformer in in-context learning (ICL) mode (solid line) and task vector prompting (TVP) mode (line with triangular markers) across problem dimensions $d \in \{4, 5, 6, 7, 8, 9\}$ and varying model depths $L$.} The dashed line indicates the random-guess baseline. The results indicate that smaller problem dimensions ($d = 4$ to $d = 7$) and shallower model depths ($L = 3$ to $L = 5$) yield stronger and more stable task encoding in the TVP mode. However, task encoding remains noisy in most cases. Task vectors emerge more clearly when the ICL loss plateaus but tend to disperse with increasing in-context length. In deeper models, task information appears to distribute across layers, reducing the distinctiveness of task vectors. 
    }
    \label{fig:vanilla_varying_d}
\end{figure}

In the linear regression task, we examine how model depth and context length affect the emergence of task vectors by varying the problem dimension. Instead of fixing $d=6$, we explore $d \in \{4, 5, 6, 7, 8, 9\}$, keeping the transformer’s embedding size at 64 and adjusting the model depth from $L=3$ to $L=8$. The transformer’s performance in ICL mode (solid line) and TVP mode (line with triangular markers) is shown in Figure~\ref{fig:vanilla_varying_d}, with $\Nd = 50$ for clarity.

\begin{highlight}
    \paragraph{Finding 2:} 
    \emph{In the linear regression task, task vectors are most distinct with moderate model depth and just before the in-context learning loss plateaus.}
\end{highlight}

We observe the following characteristics of task vectors:
\paragraph{Along the Model Depth:}
Tasks with smaller problem dimensions ($d = 4$ to $d = 7$) exhibit a clear pattern of specific model depths producing strong and stable task encoding, as illustrated in the plot. However, in most cases, task encoding remains noisy. Notably, as model depth increases, the clarity of task encoding diminishes. This could be due to the model’s increased capacity to approximate the least square solution directly in a single forward pass, reducing its reliance on previously calculated task encoding.
\paragraph{Across the In-Context Length:}
Task vectors become more distinct as the in-context learning loss plateaus, but they tend to disperse afterward. For example, when $d=5$ with $L=4$ or $d=7$ with $L=4, 5, 6$, task encoding is most evident around the 4-th in-context example but then disperses into random task encodings. This suggests that the model initially learns the task in a compact and focused manner; however, over time, the task information becomes distributed more broadly across the context.

\subsubsection{Task Vector Layer Localization}\label{sec:vanilla_tv_layer}
\begin{figure}[ht]
    \centering
    \includegraphics[width=0.9\linewidth]{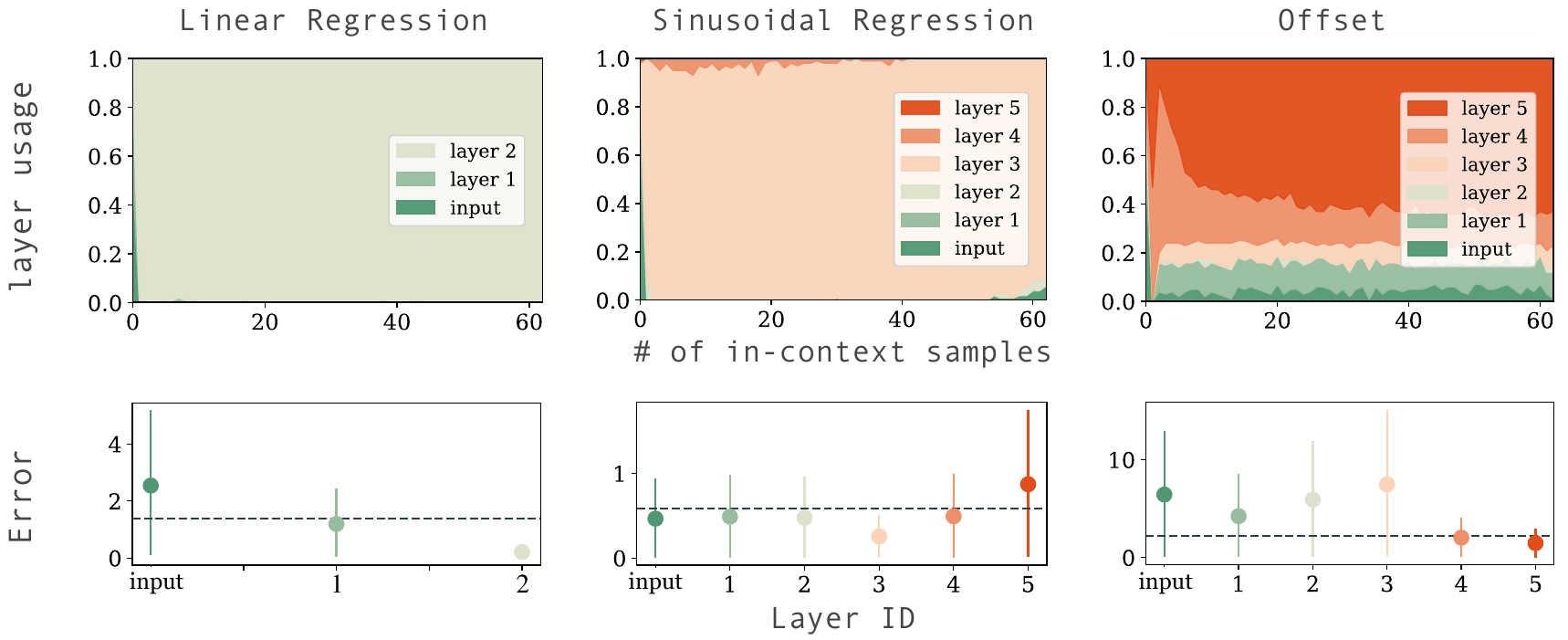}
    \caption{\small \emph{Layer selection distribution (upper row) and averaged task vector prompting performance (lower row) for task vector emergence in: Linear Regression (left), Sinusoidal Regression (middle), and Discrete Token Offset (right).} \textbf{Upper row:} we show the frequency of each layer being selected as the task vector location across varying numbers of in-context examples. For the linear regression task on a 3-layer transformer, the task vector predominantly emerges in the 2nd layer. For the more complex sinusoidal regression task on a 6-layer transformer, the task vector shifts to the 3rd layer, while for the discrete token offset task, it is primarily found in the last layer. These results suggest that task complexity influences the depth at which the task vector emerges.
    \textbf{Lower row:} we present the averaged task vector prompting (TVP) performance for each layer, measured across various context lengths. The dashed line represents random-guess performance, indicating no task information is inferred. Notably, the selected task vector layer demonstrates significantly lower loss in TVP mode compared to other layers, confirming that the emergence of the task vector is meaningful rather than due to marginal differences in loss across layers.}
    \label{fig:vanilla_tv_layers}
\end{figure}

\begin{highlight}
    \paragraph{Finding 3:} 
    \emph{Depending on the tasks, task vectors may emerge in intermediate or later layers.}
\end{highlight}
When extracting the task vector, the extractor must determine which layer the task vector resides in, based on the $\Nd$ demonstrated prompts. 
For linear regression task with the experimental setup described in Section~\ref{sec:exp_lr}, we record the proportion of each layer being chosen as the location of the task vector and present the results in Figure~\ref{fig:vanilla_tv_layers} (upper left). As suggested by the PCA visualization of activation clusters in Figure~\ref{fig:vanilla_cluster}, among the 3 layers, the 2nd layer is predominantly selected as the location where the task vector resides.

To further clarify, we also measure model's performance in the TVP mode 
when the extracted task vector is placed at each layer and average the loss across varying context lengths. These results, shown in Figure~\ref{fig:vanilla_tv_layers} (lower left), demonstrate that the selected task vector layer achieves significantly lower TVP performance compared to other layers, confirming that the emergence of the task vector is meaningful and not due to marginal differences in loss across layers. 

This observation contrasts with the phenomenon reported in pre-trained LLMs by \citet{Hendel2023InContextLC}, where task vectors for in-context learning tasks primarily emerge in the early layers out of a total of around 20 layers. In our experiments, however, we find that task vectors consistently emerge in the penultimate layer for shallow models (e.g., 3 layers). As the model depth increases (e.g., 4 or more layers), task vectors no longer emerge distinctly at any single layer, as shown in Figure~\ref{fig:vanilla_varying_d}. Instead, task information becomes distributed across multiple layers, reducing the distinctiveness of any specific layer as the task vector’s location. For additional details, Figure~\ref{fig:vanilla_tv_layers_lr} in Appendix~\ref{appendix:vanilla_tv_layer} extends this analysis to the linear regression task with $d=6$ and model depths $L=4, 5, 6$.

One factor to consider is the nature of the task at hand. For the linear function class, the ``task'' appears to be identifying the parameter $\vw$, after which the ``task execution'' phase simply requires the model to compute $\vw^T \vx$ to generate predictions. 
To explore this further, we examine the sinusoidal regression task using the same GPT-2 model configured with an embedding size of 64, 4 attention heads, and 6 layers. Additionally, we analyze the discrete token offset task with a larger embedding size of 256.

As shown in Figure~\ref{fig:vanilla_tv_layers} (middle), for this 6-layer transformer model, the 3rd layer is primarily selected as the task vector location. This finding aligns with observations in pre-trained LLMs and provides the insight that the task vector’s emergence layer in trained-from-scratch models depends on the characteristics of the task being performed. Additionally, for the discrete token offset task, the 5th layer is most frequently selected as $l^{\star}$, with other layers also being selected occasionally at lower probabilities.

\section{A New Training Algorithm to Encourage the Formation of Task Vector}\label{sec:training_algo}
As evidenced in the previous section, it is difficult to locate a single vector in the trained-from-scratch small-scale transformer model that cleanly encodes the task at hand. In this section, we propose a training algorithm that explicitly encourages the formation of task vectors (as defined in~\cite{Hendel2023InContextLC}) in the in-context learning process. 

Through this algorithm, which is a straightforward extension of the task vector definition, we obtain a model with the task vector explicitly formed. As demonstrated in Section~\ref{sec:regression_result}, comparing this model to the vanilla model (where task vectors are not explicitly formed) reveals that task vector formation enhances the model’s robustness in in-context learning tasks, particularly in scenarios with out-of-distribution prompts. This highlights the value of task vector formation in improving generalization and interpretability in in-context learning.

\subsection{A New Training Algorithm to Encourage Task Vector Formation}\label{sec:train_alg_regression}
To encourage the formation of task vectors, we include the performance metric of the task vector in the training loss. Specifically, following the notation in Section~\ref{sec:task_vector_in_vanilla}, let each random prompt be $P_k = [\vz, \vx_1, f(\vx_1), \cdots, \vx_k, f(\vx_k), \vx_{k+1}]$ to be sampled following the prompt distribution $\mathcal{P}$, and a random test prompt $[\vz, \vx_{test}]$. We train the transformer $M_{\theta}$ by optimizing the following loss: 
\begin{eqnarray}
\min_{\theta} \mathbb{E}_{P\sim\mathcal{P}} \ \sum_{i=1}^{k} \left[ \underbrace{\ell(M_{\theta}(P_i), f(\vx_{i+1}))}_{\text{ICL-loss}} + \underbrace{\ell(f(\vx_{\text{test}}), M_{\theta}([\vz, \vx_{\text{test}}]; \tv))}_{\text{TVP-loss}}\right],
\label{TVP-loss}
\end{eqnarray}
where \(\tv = h_i^l\) represents the hidden state at the $l$-th layer of the transformer, extracted from the $i$-th in-context example. 
Instead of adaptively identifying the layer where the task vector resides, we simplify the process by directly designating a specific layer $l$ as the location for forming the task vector. This ensures consistent representation of task-specific information and facilitates training. In Section~\ref{sec:regression_result}, we further explore the impact of this hyper-parameter choice on the model’s performance.

\begin{figure*}[h!]
    \centering
    \includegraphics[width=0.8\linewidth]{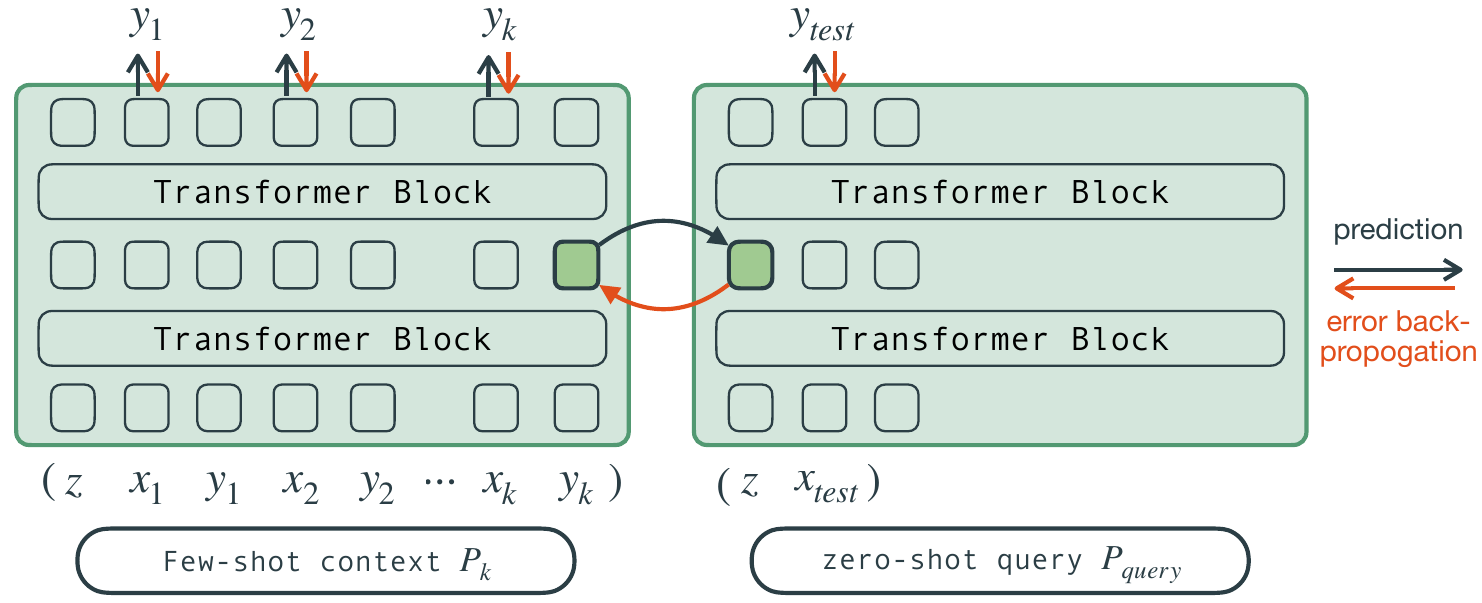}
    \caption{\small \emph{Demonstration of our training algorithm}. In vanilla Meta-ICL training, the model is updated using the ICL-loss signal from the \emph{few-shot context}. To encourage the formation of task vectors, we also explicitly include the TVP-loss from the \emph{zero-shot query}. 
    This means the model is asked to predict $y_{\text{test}}$ when only $\vx_{\text{test}}$ and the injected hidden states are given. 
    In the given illustrated example, there are in total 2 layer in the transformer model, and we set $l=1$ to encourage the formation of the task vector at the first transformer block's output.
    }
    \label{fig:method}
\end{figure*}

The loss comprises two complementary components:
\begin{inparaenum}[(1)]
    \item \textbf{In-Context Learning Loss (ICL-loss)}: The term \(\ell(M_{\theta}(P_i), f(\vx_{i+1}))\) trains the model to predict the output \(f(\vx_{i+1})\) for the $(i+1)$-th example, based on the preceding context $P_i$. Summing over $k$ examples ensures the model learns from the entire context effectively.
    \item \textbf{Task Vector Prompting Loss (TVP-loss)}: The term \(\ell(f(\vx_{\text{test}}), M_{\theta}([\vz, \vx_{\text{test}}]; \tv))\) with $\tv=h_i^l$ evaluates the model’s ability to use the injected task vector $h_i^l$, derived from the hidden state at the $i$-th in-context example, to predict the test output \(f(\vx_{\text{test}})\). This term encourages the model to encode task-specific information in the task vector $\tv$.
\end{inparaenum}
As evidence of this, in Figure~\ref{fig:overview_icl_tv}, although the model trained with ICL-loss already exhibits emergence of task vector, the model with TVP-loss further improves the encoded task vector, achieving a lower TVP error.
We illustrate the training algorithm in Figure~\ref{fig:method}, where the red arrow indicates the flow of the gradient. 

\subsection{Experimental Results on Synthetic Tasks}\label{sec:regression_result}
We apply this training algorithm to the aforementioned three synthetic tasks: (1) the linear regression task, (2) the sinusoidal regression task, and (3) the discrete token offset task described in Section~\ref{sec:exp_lr}. For all experiments, the transformer model is configured with a total of 8 layers. For consistency to the previous results, we set $\Nd=50$ here as well.

\begin{highlight}
    \paragraph{Finding 4:} 
    \emph{The proposed TVP-loss enhances task vector prompting performance, aligning it closely with in-context learning performance.}
\end{highlight}

\begin{figure}
    \centering
    \includegraphics[width=\linewidth]{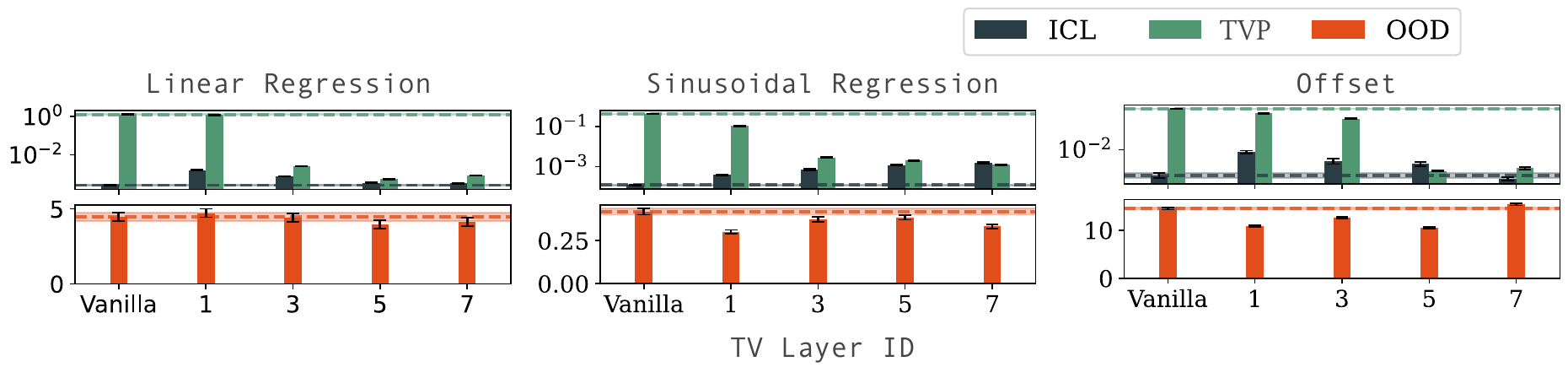}
    \caption{
    \small \emph{Model performance with vanilla and TVP-loss training.} This figure compares the performance of models trained with vanilla training and those trained with TVP-loss across different layers where task vectors are formed. The upper row shows in-context learning (ICL) and task vector prompting (TVP) performance, while the bottom row illustrates out-of-distribution (OOD) prompt performance. Dashed lines represent the performance of vanilla-trained models for reference. Each column corresponds to a specific task: linear regression (left), sinusoidal regression (middle), and discrete token offset (right). The horizontal axis indicates the layer in which the task vector is formed, which impacts TVP performance but not ICL performance.
    \textbf{ICL performance:} Models trained with TVP-loss achieve comparable ICL performance to vanilla-trained models.
    \textbf{TVP performance:} Models trained with TVP-loss exhibit improved TVP performance, particularly when task vectors are formed in intermediate or later layers of the model.
    \textbf{OOD robustness:} Models with TVP-loss demonstrate stronger generalization to OOD prompts compared to vanilla-trained models, particularly in more complex tasks like sinusoidal regression and discrete token offset, and when task vectors are formed at specific layers.
    }
    \label{fig:vanilla_vs_tv_icl}
\end{figure}

For clarity, we evaluate the in-context learning and task vector prompting performance at the 63rd context (i.e., the final query) for models trained with and without the proposed TVP-loss. These results remain consistent across various context lengths, with full details provided in Appendix~\ref{appendix:regression_full_result}. In this experiment, we designate task vectors to form at the 1st, 3rd, 5th, and 7th layers. As illustrated in Figure~\ref{fig:vanilla_vs_tv_icl}, for the 8-layer transformer, the task vector prompting performance of the vanilla-trained model is nearly random. In contrast, models trained with the TVP-loss exhibit task vector prompting performance that closely matches their in-context learning performance, as long as the task vector layer is set to an intermediate or later layer rather than the initial layers of the model.

\paragraph{Performance on OOD Prompts.} 
Forcing the task vector to form at a specific layer can be interpreted as introducing an implicit bottleneck architecture within the model’s forward pass. In this section, we examine the model’s generalization ability when the task vector is constrained to form at a particular layer. For the three synthetic tasks described earlier, we evaluate the following out-of-distribution (OOD) tasks:
\begin{enumerate}
    \item \emph{Linear Regression}: We introduce an OOD prompt using the quadratic regression task, defined as \(f(\vx_i) = \vw^\top (\vx_i \cdot \vx_i)\), where $\cdot$ denotes element-wise multiplication.
    \item \emph{Sinusoidal Regression}: we consider the OOD prompt where the task is $f(\vx_i) = \sqrt{\vw^T \vx_i}$. Considering sinusoidal regression as a composite function, with $f_1(\vx_i) = \vw^T\vx_i$, and $f_2(y) = \sin(y)$, we replace the $f_2(y) = \sin(y)$ with $f_2(y) = \sqrt{y}$ in this OOD task.
    \item \emph{Discrete Token Offset}: the OOD prompt switches the context at the 6th position.
\end{enumerate}

\begin{highlight}
    \paragraph{Finding 5:} 
    \emph{Strong task vectors improve in-context learning performance on out-of-distribution prompts.}
\end{highlight}

Additional OOD tasks and their results are presented in Appendix~\ref{appendix:more_ood}. Note that during training, none of these OOD settings were encountered during training.
As shown in Figure~\ref{fig:vanilla_vs_tv_icl}, models with task vectors formed at specific layers exhibit equal or improved performance on OOD prompts compared to vanilla-trained models. For the linear regression task, forming the task vector at the 5-th and 7-th layer results in slightly better OOD performance. 

In the sinusoidal regression task, which involves compositional functions, models with task vectors show enhanced generalization to the modified compositional task, indicating that capturing task representations at intermediate layers improves OOD generalization. This aligns with findings from~\citet{Kobayashi2024WhenCT,Schug2023DiscoveringMS}, where the use of an explicit bottleneck architecture in compositional tasks demonstrated improved compositional generalization ability. 

In the discrete token offset task, models with task vectors handle context changes significantly better than those trained without the auxiliary loss. This suggests that the auxiliary loss enables the model to adapt more effectively by filtering out unrelated context at earlier positions, compared to vanilla-trained models.

\section{In-Context Learning for Formal Language}\label{sec:task_vector_formal_language}

In this section, we extend our study to synthetic formal language settings and investigate the impact of forming task vectors within the model. We conduct experiments on the following two benchmarks: the Generative IN-Context learning dataset (GINC) introduced in \cite{Xie2021AnEO} and the RegBench dataset introduced in \cite{Akyrek2024InContextLL}.

\subsection{The GINC Dataset: the Factorial Hidden Markov Chain.}\label{sec:ginc}
\paragraph{Experimental Setup.}
The GINC dataset is introduced in~\citet{Xie2021AnEO} to study the behavior of language pretraining and the emergence of in-context learning. We follow the setup as described in~\citet{Xie2021AnEO}: we define a uniform mixture of HMMs over a family of five concepts. Each entity and property is assigned unique tokens, resulting in a total vocabulary size of 100. To generate samples, we first select an HMM from the mixture and use it to produce documents containing 576 tokens. Next, we randomly insert 192 dummy tokens, denoted as ``\texttt{-},’’ into these documents. These dummy tokens are labeled as \texttt{-100} to be ignored during processing, resulting in documents with a total length of 768 tokens.

We use the GPT-2 model with 8 layers, 8 heads, and an embedding dimension of 256. The context length of this transformer is set to 768. To train the model to form a task vector at the $l$-th layer, we sample 10 extra tokens from the same HMM and randomly insert a dummy token ``\texttt{-}''. This forms the ``zero-shot'' document. The task vector loss is then calculated by injecting the $l$-th layer's hidden states at the token ``\texttt{-}'' into the ``\texttt{-}'' token in the zero-shot document. This training pipeline (shown in Figure~\ref{fig:ginc_training}) differs from the Meta-ICL pipeline described earlier (Figure~\ref{fig:method}) because GINC specifies a pre-trained distribution that is different from the prompt distribution. In GINC, there are no explicit input-output pairs; instead, the entire trajectory performs one task. Therefore, we encode the task vector into the inserted token ``\texttt{-}'' in the trajectory.

To evaluate in-context learning performance, we follow the setup described in~\citet{Xie2021AnEO} and sample prompts from the five concepts used during training. An example of an in-context prompt with a context length of $n=3$ and example length $k=4$ is: \texttt{a b c - d / e f g - h / i j k - l}. Similarly, an example zero-shot prompt with example length $k=4$ would be: \texttt{m n o - p}. Here, the input \(\vx\) represents a short document, such as \texttt{a b c}, and the task $f$ corresponds to an underlying hidden Markov model. The output \(f(\vx)\) is the next sampled token in the document. Unlike the synthetic tasks described earlier, this function $f$ is stochastic.

To evaluate task vector prompting performance, we extract the hidden states at the i-th ``\texttt{-}'' token and inject them into the $l$-th layer activations of the zero-shot prompt at its ``\texttt{-}'' token. This setup aligns with the input format described in~\citet{Hendel2023InContextLC}. Notably, because the first ``\texttt{-}’’ token in the in-context prompt appears after an initial document of the HMM (e.g., the \texttt{a b c} document in the earlier example) and is then injected into the zero-shot prompt, additional context information from this HMM document may be included, which could lead to better task vector prompting performance compared to in-context learning performance. Through out the experiment, we set $\Nd=1$, i.e. the task vector is extracted from a single demonstrated prompt.

\begin{figure*}
    \includegraphics[width=\textwidth]{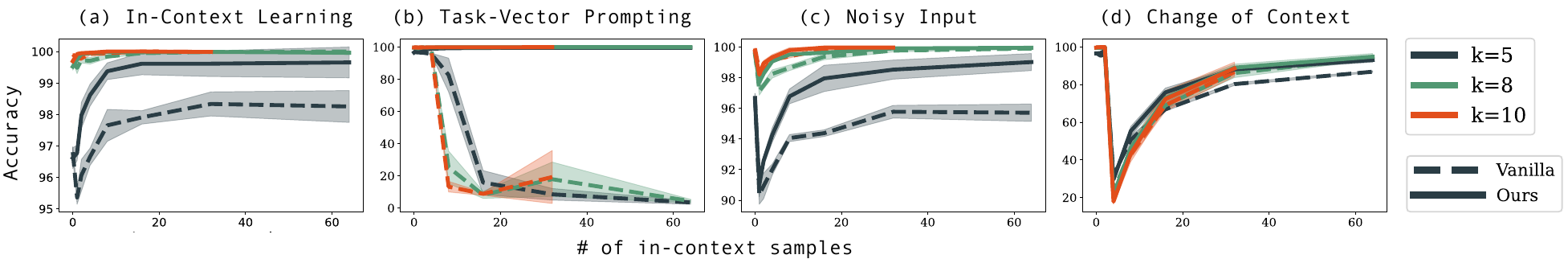}
    \caption{\small \emph{GINC dataset in-context learning performance.} Comparison of the in-context learning performance for the vanilla-trained model (dashed line) and the model trained with TVP-loss to encourage the formation of task vectors at the 5th layer (solid line). From left to right, we measure: (a) in-context learning, (b) task vector prompting, (c) in-context learning with noisy labels, and (d) in-context learning with changes in the context. In all cases, the model with task vector formation outperforms the vanilla-trained model. We follow the definition from~\cite{Xie2021AnEO} in letting $k$ to be the number of example presented in each context sample. 
    Please refer the detailed setup in Section~\ref{sec:ginc} and in~\citet{Xie2021AnEO}.
    }
    \vspace{-1em}
    \label{fig:ginc_icl}
\end{figure*}

\paragraph{In Context Learning Results.}
Though the model is not Meta-ICL trained, when training with the TVP-loss, during in-context learning, the task vector can still be formed. 
We present the result in Figure~\ref{fig:ginc_icl}. From left to right, we measure both model's performance on (a) in-context learning, (b) task vector prompting, (c) noisy in-context learning, where the in-context token is replaced with a random token with probability 0.1, and (d) when the underlying HMM is changed at the second context. 

Among all the in-distribution and out-of-distribution setting, model trained with TVP-loss consistently out-perform the vanilla-trained model. Specifically, the results indicate:
(a) the effectiveness of picking up in-context learning skills when the model is only trained on next-token prediction on documents;
(b) the successful formation of the task vector even when this in-context learning task is not directly optimized during pre-training;
(c) and (d) indicating the robustness to noise in the input prompt when trained with TVP-loss.

\subsection{The RegBench Dataset: the Probabilistic Finite Automata.}
\paragraph{Experimental Setup.}
The RegBench dataset is introduced in~\citet{Akyrek2024InContextLL} to study the behavior of language pretraining and the emergence of n-gram head. We follow and revise the setup of~\citet{Akyrek2024InContextLL}: we sample 100 deterministic finite automata (DFAs) for training.
At each training iteration, we randomly generate 5,000 training sequences from these DFAs.
Each DFA has a fixed initial state, and each state transits to another state given an input token.
For each DFA, the total vocabulary size (i.e., the number of unique tokens) is set to 40, the maximum number of states is set to 20, and each state is allowed a maximum of 10 outgoing edges.
To generate samples (sequences) from a DFA, we convert it to a probabilistic finite automaton (PFA) by assigning uniform transition probabilities across all outgoing edges, following~\citet{Akyrek2024InContextLL}.

Each prompt contains 32 samples (sequences) generated by the sampled DFA and those example are with length $k$ separated by the special token ``\texttt{|}''.
To help the formation of task vectors, we insert a dummy token ``\texttt{>}'' before the last token of the example.
After inserting \texttt{>}, a prompt with context length $n=3$ and example length $k=4$ would be like ``\texttt{abc>d|efg>h|ijk>l}''.

We use the GPT-2 model with 8 layers, 2 heads, and an embedding dimension of 128 following~\citet{Akyrek2024InContextLL}.
For each training sequence, we sample an additional example from the same DFA and insert a dummy token \texttt{>} before the last token of the example. This serves as a zero-shot prompt.
Similar to the GINC dataset, the input  $\vx$ corresponds to the generated sequence of tokens from the PFA.
The task  $f$ is determined by the PFA’s transition function, and the output  $f(\vx)$ is the next token sampled by the DFA. Importantly, this transition function is stochastic as well.
During training, we use the next token prediction loss on all tokens except the special tokens ``\texttt{>}'' and ``\texttt{|}''.
During inference, we evaluate task vector prompting performance with \(\Nd=1\).

\begin{figure*}
    \includegraphics[width=\textwidth]{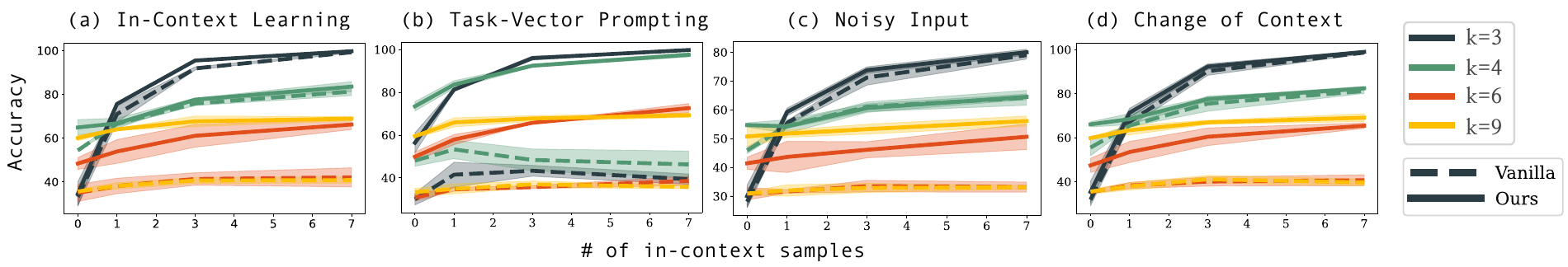}
    \caption{\small \emph{RegBench dataset in-context learning performance.} Comparison of the in-context learning performance for the vanilla-trained model (solid line) and the model trained with TVP-loss at the 4th layer (dashed line). From left to right, we measure: (a) in-context learning, (b) task vector prompting, (c) in-context learning with noisy labels, and (d) in-context learning with changes in the context. In all cases, the model with TVP-loss outperforms the vanilla-trained model. 
    }
    \vspace{-1em}
    \label{fig:regbench_icl}
\end{figure*}

\paragraph{In Context Learning Results.}
We test with various $k=3,4,6,9$, and present the results in Figure~\ref{fig:regbench_icl}, which include: (a) in-context learning (ICL) performance, (b) task vector prompting (TVP) performance, (c) noisy in-context learning, where 4 out of 32 in-context labels are replaced with random tokens, and (d) performance when the underlying DFA of 4 out of 32 samples is changed to another DFA.

For evaluating the next predicted token in the in-context learning prompt, we follow the same setup as in~\citet{Akyrek2024InContextLL}: as long as the next predicted token belongs to the outgoing edge set of the current DFA state, it is considered a correct prediction; otherwise, it is incorrect. Intuitively, as $k$ increases, more transitions are presented to the transformer. This requires the model to approximate a larger $n$-gram distribution, leading to a degradation in performance.

As shown in the Figure~\ref{fig:regbench_icl}, models trained with TVP-loss show a slight improvement in ICL performance (as shown in (a)) and a significant improvement in TVP performance (b). Furthermore, (c) and (d) demonstrate that models trained with task vectors are more robust to label noise and out-of-distribution (OOD) in-context samples. Specifically, these models maintain better performance compared to vanilla training when in-context examples include label noise or OOD samples.

\section{Discussion and Conclusions}
\paragraph{\textbf{Factors Affecting Task Vector Emergence.}}
The emergence of the task vector in  pre-trained large language models may be attributed to multiple factors, such as the model's capacity, scaling laws, and the diverse tasks encountered during pre-training. While \citet{Hendel2023InContextLC} and related studies identify and analyze the existence of task vectors, they do not study the factors that may effect their emergence and performance. This gap in understanding motivates our investigation: 
we reproduce this phenomenon in the small-scale setting when trained from scratch, gaining insight into the factors influencing task vector emergence, such as input format, model depth, and context length. For instance, in the linear regression task, when the model is trained with prompts where $x$ and $y$ alternate closely, a model with moderate depth encourages the emergence of task vectors, which is most evident before the in-context learning loss plateaus.

\paragraph{\textbf{Training with TVP-Loss.}} While task vectors can emerge through normal training processes, adding the TVP-loss to the training process encourages the formation of task vectors at prescribed locations and leads to improved accuracy and robustness. 
Across various benchmark datasets, we have demonstrated that prescribing the task vector to an intermediate or later layer of the model results in comparable in-context learning performance while forming strong task vectors. Depending on the task, task vectors formed at specific layers can enhance the model’s robustness to out-of-distribution (OOD) prompts, most likely at the intermediate or later layer.

\paragraph{\textbf{Potential Applications of Task Vectors}} 
 Task vectors can serve as a soft prompt, effectively compressing the entire context into a single vector representation. In practical scenarios, where the number and nature of tasks are unknown during pre-training and only demonstrations are provided to the model, task vectors enable both context summarization and task identification.
 Specifically, a model with task vectors formed gains the ability to identify the underlying task from demonstrations, cluster tasks using the extracted task vectors, and subsequently perform zero-shot inference with these extracted vectors.
 
 Another practical benefit is during inference, attention computations over earlier tokens can be masked out after the prescribed $l$-th layer, restricting attention to only the task vector and the query. This approach mirrors the findings of \citet{Sia2024WhereDI}, who demonstrated a 45\% reduction in computation for pre-trained LLMs. 
Looking ahead, incorporating the TVP-loss as an auxiliary objective during training offers a promising approach to enhance task-specific representation and overall performance.

\section{Acknowledgement}
The work of Robert Nowak is supported in part by NSF awards DMS-2134140, CIF-2427440, and ONR MURI N00014-20-1-2787. The work of Dimitris Papailiopoulos is supported in part by ONR Grant No. N00014-21-1-2806 and No. N00014-23-1-2848. The work of Kangwook Lee is supported in part by NSF CAREER Award CCF-2339978, Amazon Research Award, and a grant from FuriosaAI.

\bibliography{ref}
\bibliographystyle{abbrvnat}

\newpage
\appendix
\onecolumn

\renewcommand{\contentsname}{Organization of the Appendix}
\addtocontents{toc}{\protect\setcounter{tocdepth}{3}}
\tableofcontents

\clearpage
\section{Related Works}

\paragraph{Understanding In-Context Learning Behavior in the Pre-trained Large Language Model.}
Since the launch of GPT-3~\citep{Brown2020LanguageMA} and the observation that, during inference, the model can leverage few-shot examples to improve performance, researchers have been keenly interested in understanding the mechanisms of this in-context learning behavior. Several studies have explored the significance of labeling in context demonstrations~\citep{Min2022RethinkingTR,Kim2022GroundTruthLM,Kossen2023InContextLL, Long2024DecomposingLS}, while others have examined the phenomenon through circuit mechanisms~\citep{elhage2021mathematical,Wang2022InterpretabilityIT,Singh2024WhatNT,Hanna2024HaveFI,Wang2023LabelWA}. Researchers have also attempted to decode the logits to reveal how transformers refine their output during the forwarding process~\citep{nostalgebraist,dar2022analyzing,geva2022transformer}. Research by \cite{Dai2022WhyCG}, \cite{Geva2022TransformerFL}, and \cite{merullo2023mechanism} links in-context learning behavior to implicit internal weight and activation updates within the model. To better understand ``learning'' in-context, several works~\citep{Pan2023WhatIL, Sia2024WhereDI} have studied its two distinct phases--task recognition and task learning--using controlled experiments designed to disentangle these phases.
These studies focus on pre-trained large language models without access to the pre-training procedure. The literature on embedding editing and task vector will be discussed in detail later.

\paragraph{Understanding In-Context Learning Behavior in the Controlled Small-Scale Setting on Data-Fitting Problems.}
Researchers have also examined in-context learning behavior in small-scale models, where models are trained from scratch to perform in-context learning tasks. \citet{Garg2022WhatCT,Akyrek2022WhatLA,Oswald2022TransformersLI,Oswald2023UncoveringMA} investigated the ability of transformers to learn regression problems, interpreting the model as performing a single step of gradient descent. This line of research was extended further in different pre-training mixtures of tasks~\citep{raventos2023effects, Yadlowsky2023PretrainingDM}, the ability to generalize to unseen tasks~\cite{yadlowsky2023can}, explanations via the Bayesian optimal estimator~\citep{Zhang2023WhatAH,Bai2023TransformersAS,Muller2021TransformersCD}, the viewpoint of generalization error~\citep{Li2023TransformersAA}, and the perspectives of task retrieval and task learning~\citep{lin2024dual,NEURIPS2023_73950f0e}.

Beyond regression, several studies have explored in-context learning abilities in other domains, such as reinforcement learning~\citep{Lee2023SupervisedPC, Lin2023TransformersAD}, discrete function learning~\citep{Bhattamishra2023UnderstandingIL}, factorial hidden Markov chains~\citep{Xie2021AnEO}, and deterministic finite automata (DFAs)~\citep{Akyrek2024InContextLL}. Additionally, \cite{Chan2022TransformersGD, Chan2022DataDP} investigated the behavior of in-context learning and in-weight learning on the Omniglot datasets.

\paragraph{Task Representation}
The notion of a "task vector" was first introduced by \cite{Ilharco2022EditingMW}. Subsequently, \citet{Hendel2023InContextLC}, \citet{Merullo2023LanguageMI}, \citet{Yu2023CharacterizingMF}, \citet{Liu2023IncontextVM}, \citet{saglam2024learning} and \citet{Li2024ImplicitIL} demonstrated that a single vector in the model's activation space can encode learned functions in a pretrained model. Specifically, \cite{Liu2023IncontextVM} showed that, when presented with an in-context learning demonstration, the activation at each layer points to a subspace encoding the task information. Additionally, \cite{Todd2023FunctionVI} identified the "function vector," another form of task vector, by averaging the causal attention heads, which can also guide the pre-trained language model's performance towards desired tasks. Task vectors have also been identified under different modality~\citep{Luo2024TaskVA, Hojel2024FindingVT, Peng2024LearnableIV}, and from different perspective, such as cognitive science~\citep{Piantadosi2024WhyCA}, and through theory perspective~\citep{Tao2024TaskAT}. 

A recent study by~\citet{Mittal2024DoesLT} investigates a setting similar to ours, focusing on whether learning appropriate latent representations—achieved through a bottleneck architecture—can enhance robustness in in-context learning. In contrast, our approach facilitates the natural emergence of latent representations within the model’s architecture, aligning with the phenomenon of task vector emergence. Building on this idea,~\citet{Kobayashi2024WhenCT} further examined bottleneck architectures, demonstrating that models incorporating bottlenecks exhibit improved compositional generalization.~\citet{elmoznino2024context} use the bottleneck architecture to examine the prequential ICL performance.~\citet{Han2024EmergenceOA} investigate the training dynamics of task vector formation, demonstrating that as task vectors emerge, the model simultaneously develops conditional decoding algorithms, thereby enhancing its in-context learning (ICL) performance.

\paragraph{Task Vector Applications}
\cite{Li2024InContextLS} utilized this task vector to enhance test-time adaptation, while \cite{Pham2024RobustCE} employed it to erase malicious concepts during pre-training. Beyond activations, task information can also be encoded in a task token~\cite{Bai2024IdentifyingAA}, or a pause token can be used to gain extra computation time~\cite{Goyal2023ThinkBY}. \citet{Zhao2024BeyondSC} propose to present each task concept with a Gaussian distribution. Following the notion of ``task vector'' in the weight space, several works utilize the weight space difference for chatbot on new language~\citep{Huang2023ChatVA,Kang2024WhisperMD}, emotion transfer~\citep{Kalyan2024EmotionAE}, aesthetic assessment~\citep{Yun2024ScalingUP}, and soft prompt initialization~\citep{Belanec2024TaskPV}.

\paragraph{Soft Prompt and Efficient Adaptation of Transformers.}

In this work, we primarily follow the definition of a task vector as described by \cite{Hendel2023InContextLC}. This task vector effectively compresses the information in the context into a single vector, which can be injected into the model to perform zero-shot inference. The notion of a task vector is also related to the concept of a soft prompt~\citep{Lester2021ThePO,He2022HyperPromptPT,Liu2021GPTUT,Xu2023CompressTP,Kang2024WhisperMD}, where the context information is similarly compressed. Beyond learning soft prompts that encode instructions, researchers have also explored context compression by modifying the attention mask to ensure the summarization of context at a specific token~\cite{Ren2023ContextCF,Mu2023LearningTC,Phang2024InvestigatingTE}. In the format of ICL, the researcher also study the continuous representation in context~\citep{Zhuang2024VectorICLIL}, as a reminiscent of the soft encoding.

Although we propose a training algorithm to encourage the formation of task vectors, our primary goal is not to compare our method of encoding task vectors with these approaches in soft prompt learning. Instead, we aim to understand the benefits of incorporating task vectors into in-context learning.

\section{Supplements on the Trained-from-Scratch Model}
\subsection{Effects of Input Formats}\label{appendix:prompt_format}
In this section, we examine different input formats for in-context learning prompts and their effects on the emergence of task encoding. The various prompt formats are detailed in Table~\ref{tab:input_format}. For clarity, we refer to the token positions where task vector presence is examined as the \emph{task token locations}.

We focus on the case where the problem dimension is $d=6$ and the model depth is $L=3$, as it exhibits the most pronounced task encoding, as shown in Figure~\ref{fig:vanilla_varying_d}. In Figure~\ref{fig:input_format_extractor}, we evaluate various task vector extraction methods (detailed in Appendix~\ref{appendix:tv_extractor}), with their respective task vector prompting performance shown as line with triangular markers. Notably, only the model with the input format \texttt{(*xy)} demonstrates task encoding that surpasses random baselines. By default, we use the format \texttt{(*xy)} throughout the main paper unless specified otherwise.

In the pre-trained LLM~\citep{Hendel2023InContextLC}, the input format uses \texttt{x->y}. There is a clear difference between the pre-trained LLM and the trained-from-scratch transformer in how task information is encoded. In the pre-trained LLM, task information is primarily stored in the “maps-to” token (\texttt{->}). Conversely, in the trained-from-scratch transformer, task information is stored in the $y$ token (i.e., the “label” token), but only when x and y alternate closely in the input format. We further analyze the attention maps and PCA of the activations when input format is \texttt{x->y}, following the same setup as in Figure~\ref{fig:vanilla_cluster}, and present the results in Figure~\ref{fig:x_y_attn_cluster}. Across different layers, the attention maps do not exhibit meaningful task information encoded in the patterns. Consequently, we showcase only one example attention map to illustrate the absence of distinct task encoding in this setup. Additionally, the PCA of the activations reveals that the encoded information is nearly identical across tasks.

We hypothesize that this discrepancy arises because pre-trained LLMs learn the semantic meaning of the “maps-to” symbol (\texttt{->}) from their extensive pre-training corpus, enabling them to use \texttt{->} as an anchor for task summarization. Additionally, this delimiter \texttt{->} is needed to separate the $x$ and $y$ entries. In contrast, for the trained-from-scratch transformer, the \texttt{->} token functions more like a \texttt{<pause>} token~\citep{Goyal2023ThinkBY}, providing additional computational resources rather than semantic significance. Moreover, the training procedure ensures that $x$ and $y$ tokens occupy fixed positions, removing the need for a delimiter to separate them.

Furthermore, \citet{Wang2023LabelWA} show that label words themselves can serve as anchors for aggregating task information in context. This observation aligns with our finding that task information is encoded in the $y$ tokens. While this result is not explicitly framed within the task vector framework, it highlights the critical role of label tokens in task encoding.

\begin{table}[th]
\centering
\caption{\small \emph{Different prompt formats examined.} We analyze the input format based on the settings described in~\citet{Garg2022WhatCT} and~\citet{Hendel2023InContextLC}. Here, $\vz$ represents a special token, $P_{\text{query}}$ denotes the test query format, and $P_k$ refers to the $k$-shot context. We examine the activations at $\Lambda_f$, referred to as the task token location.}
\label{tab:input_format}
\begin{center}
\resizebox{\textwidth}{!}{
\begin{tabular}{llll}
\toprule
 &  $P_{\text{query}}$ & $k$-shot examples $P_k$ & Task Token Location $\Lambda_f$ \\
\midrule
\texttt{(*xy)}~\citep{Garg2022WhatCT} & $[\vz, \vx_{\text{test}}]$ & $[\vz, \vx_1, f(\vx_1), \cdots, \vx_k, f(\vx_k), \vx_{k+1}]$ &  $P_k$\texttt{[0::2]}: $\{f(\vx_i)\}$ and $\vz$\\
\texttt{(x->y)}~\citet{Hendel2023InContextLC} & $[\vx_{\text{test}}, \vz]$ & $[\vx_1, \vz, f(\vx_1), \cdots, \vx_k, \vz, f(\vx_k), \vx_{k+1}, \vz]$ &  $P_k$\texttt{[1::3]}: $\vz$\\
\texttt{(*x->y)}~\citet{Hendel2023InContextLC} & $[\vz, \vx_{\text{test}}, \vz]$ & $[\vz, \vx_1, \vz, f(\vx_1), \cdots, \vx_k, \vz, f(\vx_k), \vx_{k+1}, \vz]$ & $P_k$\texttt{[0::3]}: $\{f(\vx_i)\}$ and $\vz$\\
\bottomrule
\end{tabular}}
\end{center}
\end{table}

\begin{figure}
    \centering
    \includegraphics[width=0.95\linewidth]{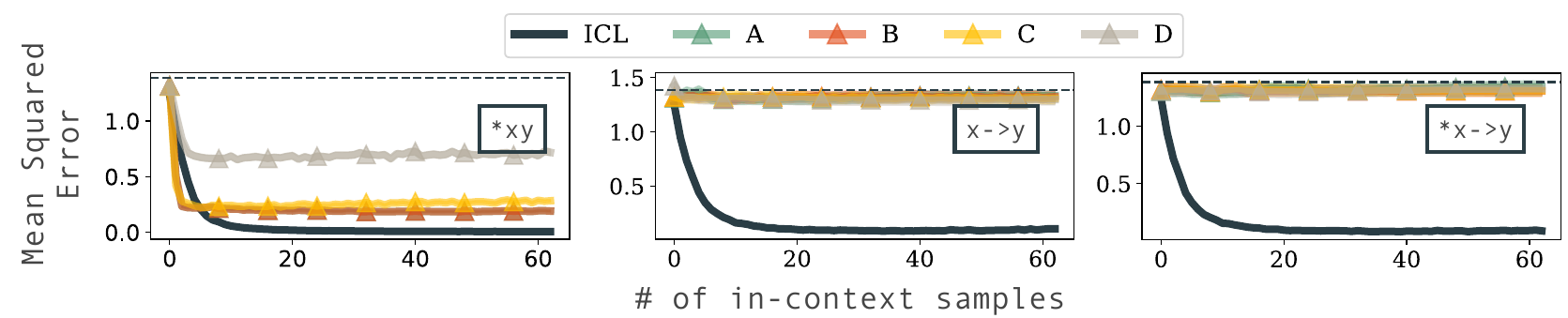}
    \caption{\small \emph{In-context learning (ICL) and task vector prompting (TVP) performance across input formats and task vector extraction methods.} This figure compares task vector prompting performance across four task vector extraction methods (detailed in Appendix~\ref{appendix:tv_extractor}) and three input formats: \texttt{(*xy)}, \texttt{(x->y)}, and \texttt{(*x->y)}. Method (A) is the default task vector extraction method, and \texttt{(*xy)} is the default input format used throughout the main paper. As shown, for the \texttt{(*xy)} input format, noticeable task encoding is observed, with performance exceeding random predictions, regardless of the extraction method. In contrast, for the \texttt{(x->y)} and \texttt{(*x->y)} input formats, no task encoding can be reliably extracted by any method, resulting in task vector prompting performance that is nearly random.}
    \label{fig:input_format_extractor}
\end{figure}

\begin{figure}
    \centering
    \includegraphics[width=0.95\linewidth]{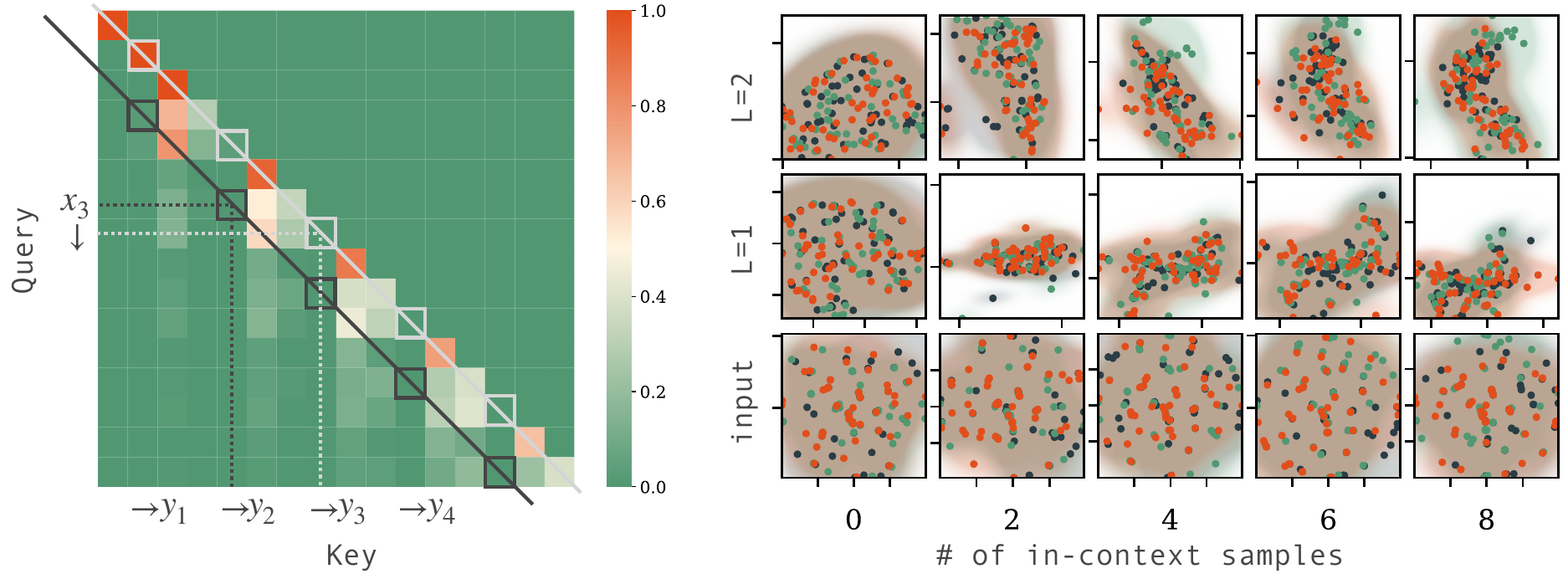}
    \caption{\small \emph{Attention map and PCA visualization of activations for three linear functions with input format \texttt{(x->y)}.}
    Following the same setup as in Figure~\ref{fig:vanilla_cluster}, we plot the attention map and PCA visualization of the \texttt{->} token’s activations. (\emph{Left}) The attention map at the 2nd layer highlights interactions between task tokens (white boxes) and the attention from $\vx_i$ tokens to task tokens (black boxes). Unlike the format \texttt{(*xy)}, this format does not exhibit a meaningful attention pattern where $\vx_i$ tokens attend to previous task tokens, or task tokens attend to prior task tokens. (\emph{Right}) PCA visualizations show that the activations of the \texttt{->} token encode nearly identical information across different context lengths and different tasks, suggesting the absence of distinct task encoding dynamics in this input format.}
    \label{fig:x->y_attn_cluster}
\end{figure}

\subsection{Various Task Vector Locating Methods}\label{appendix:tv_extractor}
In this section, we analyze different methods for extracting task vectors. The main question we address is: \emph{If task encoding is detectable by one task vector extraction method, will it also be detectable by another method?} Our focus is not on the quality of the extracted task vector but rather on whether the method can detect task encoding that is non-random. Below, we present task vector extraction methods inspired by~\citet{Hendel2023InContextLC, Liu2023IncontextVM, Li2024ImplicitIL}, which are representative approaches in the task vector literature.

Following the definition in Section~\ref{sec:def}, we study the following task vector extractor methods:
\paragraph{(A) Task Token's Output Embedding} \citet{Hendel2023InContextLC} have observed that in pre-trained language models, the $l$-th layer output embedding of the $k$-th task token, denoted as $h_k^l$, encodes the task vector. We follow this setup and define the task vector extractor $g$ as follows: Let $\tv = h_k^{l}$, then
$$
M(P_{\text{query}}; h_k^{l}) \coloneq \texttt{replace task token's l-th layer embedding with } h_k^{l}.
$$
This notion of the task vector indicates the independence of the encoded task information from the demonstrated prompt.

\paragraph{(B) Task Token's Output Embedding Difference} \citet{Liu2023IncontextVM} pointed out that when inputting two sequences $x$ and $y$, the difference in the hidden space corresponds to the mapping from $x$ to $y$ in the pre-trained model. Though their setting is slightly different from the in-context learning format, we leverage this idea and propose the following way to extract the task embedding: Following the definition of $h_k^l$ from above, and let $\tilde{h}_k^{l}$ be the hidden state of the model when inputting the test prompt $P_{\text{query}}$, then $\tv = h_k^{l} - \tilde{h}_k^{l} = \Delta h_k^{l}$, indicating the difference between the $k$-th context embedding and an uninformative context, then
$$
M(P_{\text{query}}; \Delta h_k^{l}) \coloneq \texttt{add } \Delta h_k^{l} \texttt{ to task token's l-th layer embedding.}
$$
This notion of the task vector loosens the constraint of absolute independence of the task from the demonstrated prompt.

\paragraph{(C) Principal Direction of Task Token's Output Embedding Difference} \citet{Liu2023IncontextVM} also observed that taking the principal direction along the difference of hidden states aligns more closely with the task. Motivated by their observation and setting, we propose the following task vector extractor: Following the definition of $\Delta h_k^{l}$ from above, we set $\tv = \texttt{PCA}( \Delta h_k^{l} )$, then
$$
M(P_{\text{query}}; \texttt{PCA}(\Delta h_k^{l})) \coloneq \texttt{add } \texttt{PCA}(\Delta h_k^{l}) \texttt{ to task token's l-th layer embedding.} %
$$
We also re-scale the updated hidden states to preserve the model's original capabilities. This approach differs slightly from the original setup in~\citet{Liu2023IncontextVM}, where PCA is applied to the concatenated hidden states across all layers. Their method is specifically designed for alignment tasks in natural language settings, where the context $x_i$ and $y_i$ are input separately. In our case, as the model is trained in an in-context learning (ICL) format, we adhere to the input prompt structure where $x_i$ and $y_i$ are provided alternately in one prompt. We also attempted to follow their setup by applying PCA to the concatenated hidden states across all layers; however, this approach resulted in near-random task vector prompting performance.

\paragraph{(D) Linear Combination of Task Token's Output Embedding to the Query Output Embedding} \citet{Li2024ImplicitIL} propose to learn the coefficient of the linear combination between the task token's output embedding to the query output embedding. Motivated by this, we let $\tv = \{h_k^l\}_{l \in [L]}$, then
\begin{align*}
M(P_{\text{query}}; \{h_k^l\}_{l \in [L]}) \coloneq \texttt{replace task token's l-th layer embedding with }\\ \alpha_l^h  h_k^l + \beta_l^h h_t^l \quad \forall l \in [L].    
\end{align*}
where $h_t^l$ is the output embedding at $l$-th layer for the test query, and $\alpha_l^h$, and $\beta_l^h$ are learnable parameters. To learn these coefficients, we use a constant learning rate of 0.01, the AdamW optimizer, and train for a total of 100 epochs. Following the initialization setup in~\citet{Li2024ImplicitIL}, the coefficients are initialized as $\alpha_l^h = 0.1$ and $\beta_l^h = 1$.

\subsubsection{Task Vector Extractor Performance}
We evaluate the performance of various task vector extraction methods across different input formats, as described in~\ref{appendix:prompt_format}, and present the results in Figure~\ref{fig:input_format_extractor}. When using the \texttt{*xy} input format, the task vector extraction methods demonstrate noticeable task encoding. However, for the other input formats, no informative task encoding is detected, regardless of the extraction method used.

It is worth noting that the learnable task vector extraction method (D) performs slightly worse than the other three methods, possibly due to its initialization. We did not focus on further improving this method, as the primary goal of this section is to demonstrate the detectability of task encoding rather than to optimize its quality.

\subsection{Effects of Model Capacity}\label{appendix:model_capacity}
\begin{figure}[ht]
    \centering
    \includegraphics[width=0.9\linewidth]{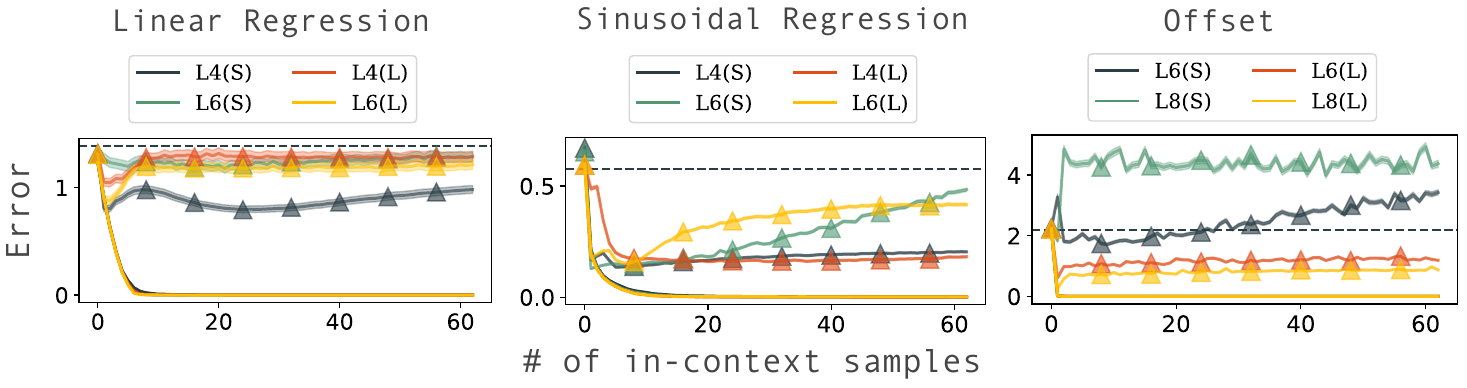}
    \caption{\small \emph{Effects of model embedding dimension on task vector prompting performance.} This figure shows the task vector prompting performance of models with different embedding sizes (64 for notation ``S'' and 256 for notation ``L''). For the regression tasks (Linear and Sinusoidal), larger embedding sizes make task encoding less prominent, likely because the increased capacity allows the model to find ``shortcut'' solutions, such as approximating the least-squares solution directly, without the need to store intermediate task information. In contrast, for the discrete token offset task, a larger embedding size enables the model to store task information in a more separable and structured manner.}
    \label{fig:vanilla_various_model_size}
\end{figure}

In this section, we examine the model with same depth, but different embedding size, to investigate the effect of model capacity. Specifically, for the three tasks mentioned in Section~\ref{sec:exp_lr}, we study the trained-from-scratch model's performance on model depth 4, 6 or 8, with embedding size 64 (with notation ``S'') or 256 (with notation ``L''), and present the result in Figure~\ref{fig:vanilla_various_model_size}.

As illustrated in the figure, increasing the embedding size allows the model to discover shortcut solutions for regression tasks. Specifically, the model approximates the least squares solution in a single forward pass, eliminating the need to store intermediate task encodings. Conversely, for the discrete token offset task, the model operates on tokens represented in a learned embedding space. To solve  $f(x) = a \times x + b$, the model must map one token to another using the task-specific parameters $a$ and $b$. We hypothesize that the nature of token embeddings makes it less straightforward for the model to learn a shortcut solution for this task. As a result, the model is compelled to store the task information explicitly. Consequently, increasing the model capacity provides more space for storing and organizing task encodings, leading to clearer task representations as capacity grows.

\subsection{Effects of Sparse Attention}\label{appendix:sparse_attention}
Sparse attention~\citep{lou2024sparser} has been proposed to improve transformer efficiency with minimal performance degradation. In in-context learning tasks (detailed in Section~\ref{sec:exp_lr}), a model could potentially solve these tasks using sparse attention: For each query, the model only needs to attend to itself and the previously stored task information. Specifically, each $x_i$ token needs to attend to two tokens ($x_i$ and $y_{i-1}$), while each $y_i$ token needs to summarize the task information by attending to three tokens ($y_i$, $y_{i-1}$, and $x_i$).

\begin{figure}
    \centering
    \includegraphics[width=0.9\linewidth]{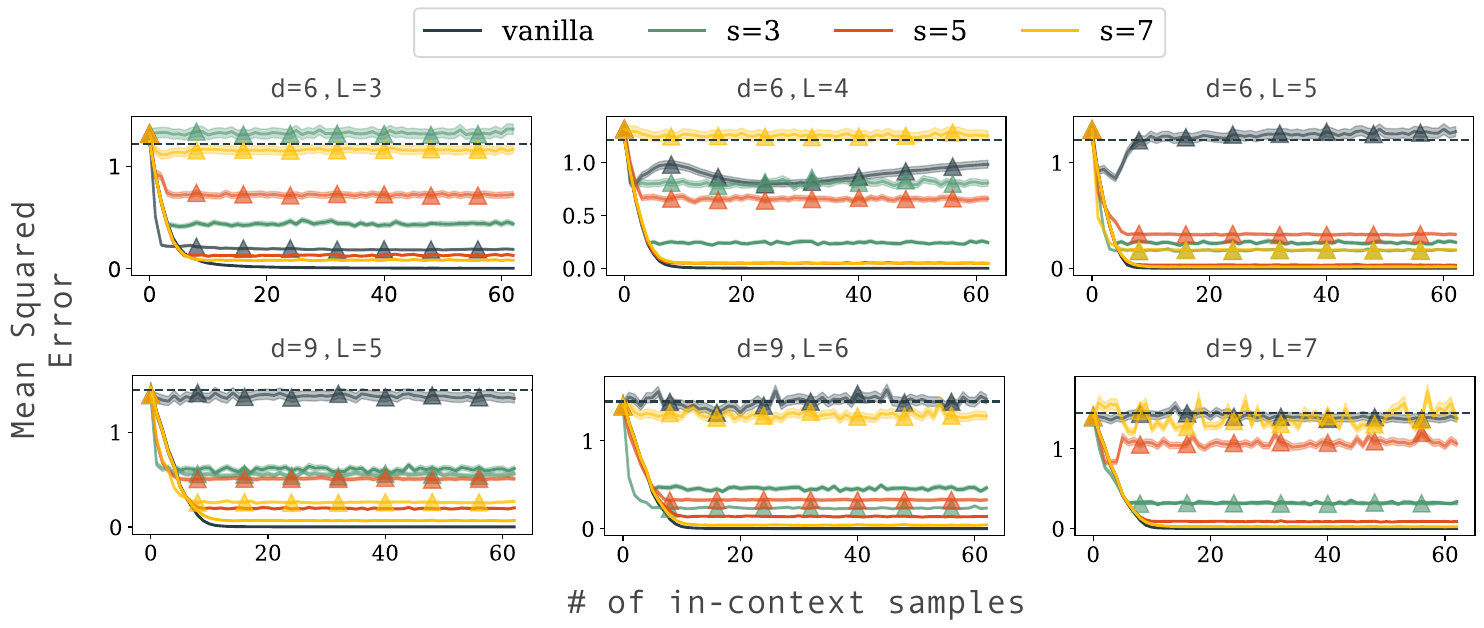}
    \caption{\small \emph{Task vector prompting performance for models trained with sliding window sparse attention.} We compare models trained with full-window causal attention (vanilla) and sparse attention using sliding window sizes of 3, 5, and 7. Solid lines represent the in-context learning performance, while lines with extra triangular marker indicate the task vector prompting performance. As shown, sparse attention enhances task vector emergence but leads to a slight degradation in in-context learning performance.}
    \label{fig:sliding_window}
\end{figure}

In this section, we investigate the impact of explicitly enforcing sparse attention constraints on in-context learning and task vector prompting performance. Using the setup described in Section~\ref{sec:exp_lr}, we apply a sliding window strategy~\citep{qiu2019blockwise, beltagy2020longformer} combined with causal attention, with window sizes $s$ of 3, 5, and 7. We evaluate the transformer with various depth on the linear regression task with $d=6$ (where task vectors emerge) and $d=9$ (where task vectors are nearly absent), and present the results in Figure~\ref{fig:sliding_window}.

As shown in the figure, models trained with sparse attention demonstrate better task encoding compared to those with full-window causal attention. For instance, when $d=6$ and $L=5$, task encoding in the full-window model fades after the first few in-context examples. In contrast, models trained with sliding window attention (window sizes 3, 5, and 7) maintain stable task encoding across varying context lengths.

However, while the sliding window improves task vector emergence, it slightly degrades in-context learning performance because each query has access to fewer tokens during training. Moreover, task encoding stability is not always guaranteed: larger window sizes improve in-context learning performance but degrade task vector prompting performance.

\subsection{Supplements on Task Vector Layer Localization}\label{appendix:vanilla_tv_layer}
To further support the argument that task vectors in the linear regression task emerge only in the penultimate layer, we extend the analysis from Figure~\ref{fig:vanilla_tv_layers} to include the linear regression task with $d=6$ and model depths $L=4, 5, 6$. As shown in the figure, for deeper models, particularly with $L=5$ and $L=6$, no single layer consistently encodes task information, and the task vector prompting performance appears random. From this analysis, we conclude that in the linear regression task, and under the experimental setup we examined, task vectors either emerge in the penultimate layer when the model is shallow or fail to emerge entirely when the model is deep.
\begin{figure}
    \centering
    \includegraphics[width=0.95\linewidth]{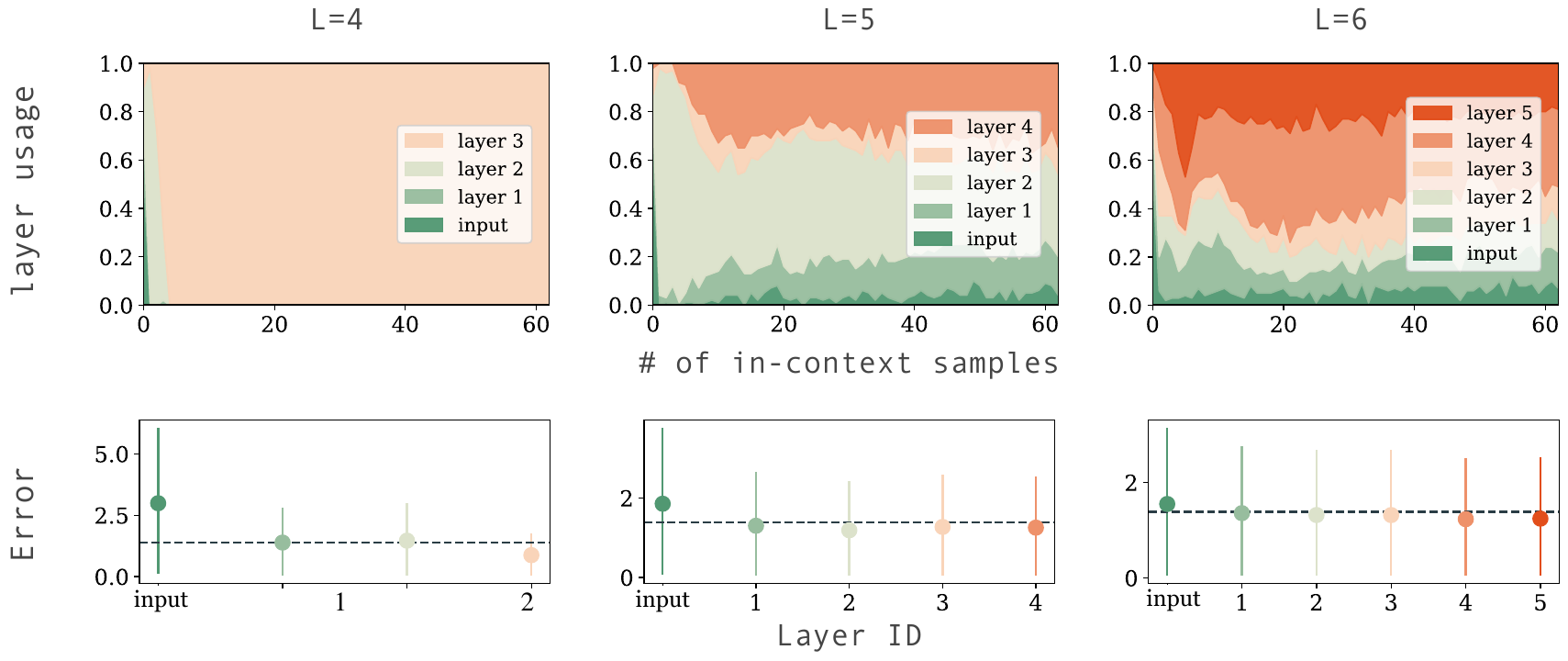}
    \caption{\small \emph{Layer selection distribution (upper row) and averaged task vector prompting (TVP) performance (lower row) for task vector emergence in Linear Regression with problem dimension $d=6$ and model depths $L=4$ (left), $L=5$ (middle), and $L=6$ (right).}
    \textbf{Upper row:} The frequency of each layer being selected as the task vector location across varying numbers of in-context examples is shown. As model depth increases, task information becomes less concentrated in any single layer, resulting in a more distributed selection of task vector locations.
    \textbf{Lower row:} The averaged TVP performance for each layer, measured across various context lengths, is presented. The dashed line indicates random-guess performance, representing a scenario where no task information is inferred.    
    These results demonstrate that for deeper models (especially $L=5, 6$), no single layer consistently encodes significant task information, and task vector performance approaches random, highlighting the distributed nature of task encoding in deeper architectures.}
    \label{fig:vanilla_tv_layers_lr}
\end{figure}

\subsection{Effects of Number of Demonstrated Prompts $\Nd$}\label{appendix:vanilla_varying_N}
\begin{figure}
    \centering
    \includegraphics[width=0.8\linewidth]{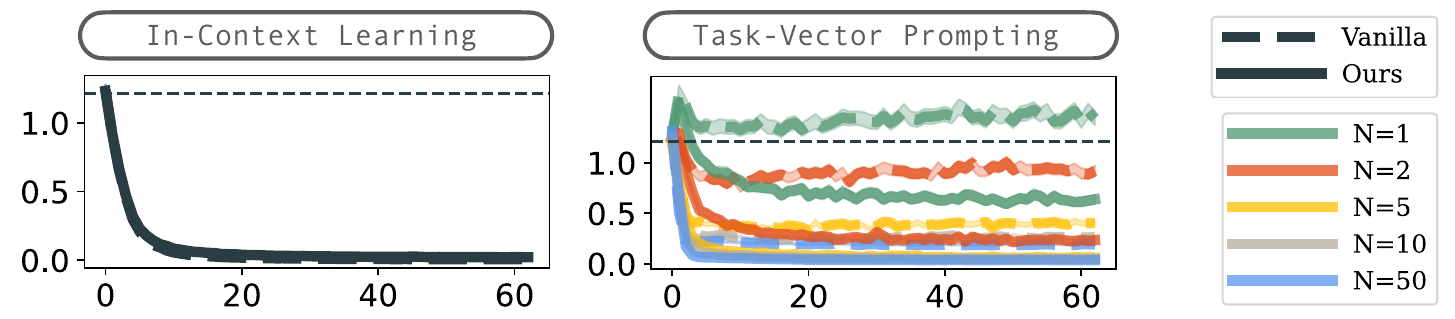}
    \caption{\small \emph{Impact of the Number of Demonstrated Prompts (\(\Nd\)) on Task Vector Prompting Performance.} Increasing \(\Nd\) improves task vector prompting performance, as larger \(\Nd\) values yield more concentrated task vectors. Across all \(\Nd\) settings, model trained with TVP-loss (“Ours”) consistently outperforms the vanilla model, demonstrating better task vector encoding. The dashed line represents random-guess performance.
    }
    \label{fig:vanilla_varying_N}
\end{figure}

Throughout the paper, we use \(\Nd=50\) by default when evaluating task vector prompting performance. Intuitively, larger \(\Nd\) values yield more concentrated and purified task vectors. To illustrate the effect of \(\Nd\), we evaluate task vector prompting performance for \(\Nd=1, 2, 5, 10, 50\), following the setup in Section~\ref{sec:exp_lr}, and present the results in Figure~\ref{fig:vanilla_varying_N}. In this experiment, the transformer is trained on a linear regression task with a problem dimension of $d=6$ and transformer depth $L=3$. For the model trained with TVP-loss, the task vector is explicitly formed at the 2nd layer.

As shown in the figure, when \(\Nd=1\), the vanilla model yields random task vector prompting performance. With increasing \(\Nd\), task vector performance improves steadily. Notably, the model trained with TVP-loss (denoted as “Ours” in the figure) demonstrates better task vector encoding compared to the vanilla model across all \(\Nd\) values. Because the linear functions in regression tasks have weights that exist in a continuous space and can be very close to each other, separating tasks with only a single demonstration is challenging. As a result, it requires \(\Nd=5\) to achieve task vector prompting performance comparable to its in-context learning performance. For the vanilla model, however, even with \(\Nd=50\), task vector prompting performance remains slightly worse than in-context learning performance.

\section{Supplements on Training with TVP-Loss}
\subsection{Full Results for Section~\ref{sec:regression_result}}\label{appendix:regression_full_result}
\begin{figure}
    \centering
    \includegraphics[width=0.9\linewidth]{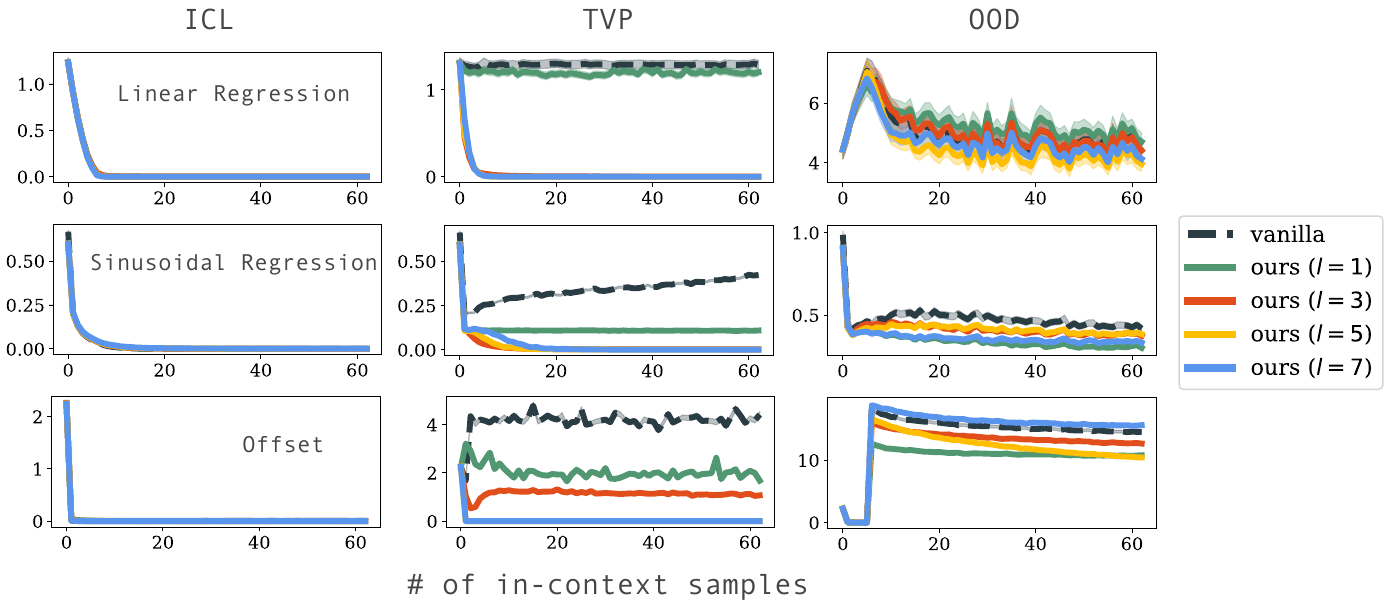}
    \caption{\small \emph{Model Performance with Vanilla and TVP-Loss Training.} We compare the in-context learning (ICL), task vector prompting (TVP), and out-of-distribution (OOD) performance of models trained with vanilla training (black) and those trained with our TVP-loss, where task vectors are formed at different layers ($l = 1, 3, 5, 7$). Full results are shown across varying numbers of in-context examples for three tasks: linear regression (top row), sinusoidal regression (middle row), and discrete token offset task (bottom row).}
    \label{fig:our_method_perf_full}
\end{figure}

In Section~\ref{sec:regression_result}, we present the in-context learning and task vector prompting performance at the 63rd context position for models trained with and without the TVP-loss, along with the in-context learning performance on out-of-distribution prompts. Figure~\ref{fig:our_method_perf_full} extends these results by showing the full performance across varying context lengths. This analysis complements the main results, demonstrating that the observations at the 63rd position are not unique but instead hold consistently across different context lengths.

\subsection{Performance on Additional OOD Tasks}\label{appendix:more_ood}
We evaluate the model’s performance on additional out-of-distribution (OOD) prompts for the regression tasks, specifically: 
(a) logistic regression, 
(b) in-context learning with an outlier context presented with a probability of 0.1, where the outlier is defined as $\vx=1$ and $y=1$.
(c) in-context learning with noisy labels (noise level 0.3), 
(d) scaled weights ($\times 3$).

For the linear regression training task, models with task vectors formed at the 7th layer (i.e., the last layer) perform equal to or slightly better than the vanilla-trained model, except when the weights and inputs are scaled. This suggests that models trained with TVP-loss exhibit reduced robustness when the input range is significantly altered. For the sinusoidal regression task, models with task vectors formed at the 3rd layer exhibit equal or improved OOD performance.

\begin{figure}
    \centering
    \includegraphics[width=0.9\linewidth]{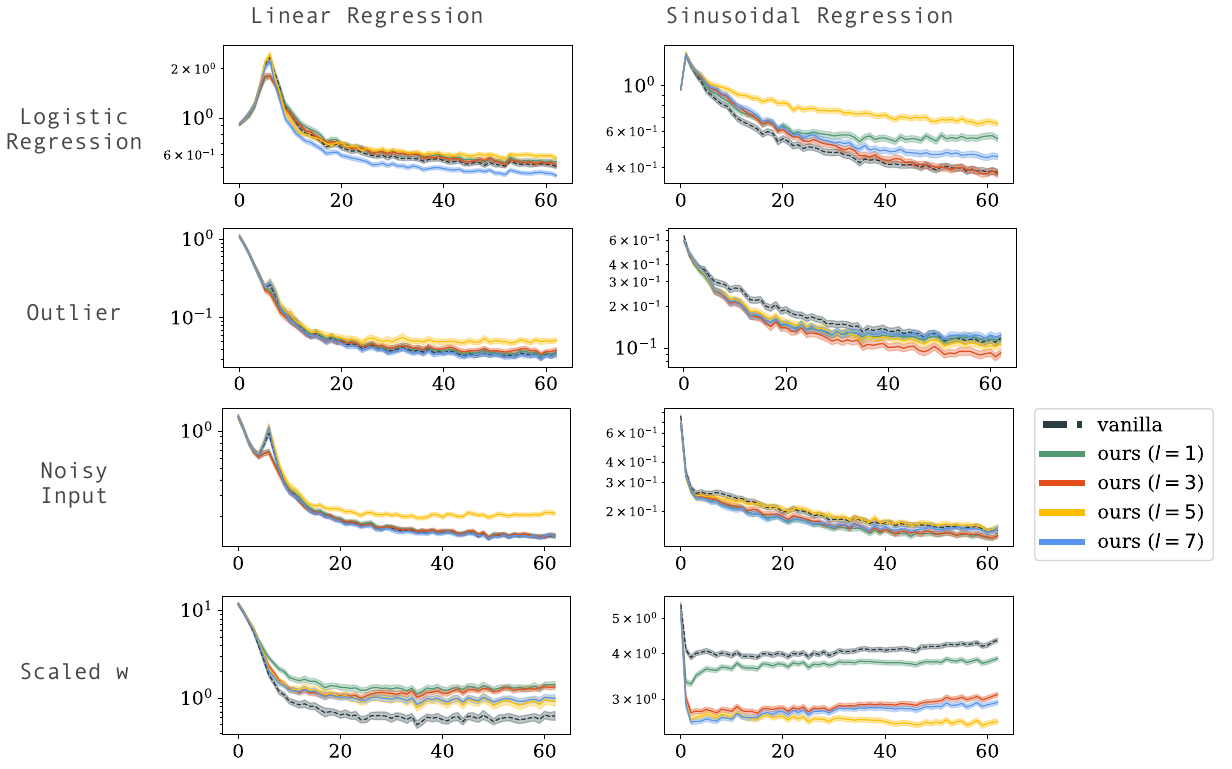}
    \caption{\small \emph{In-Context Learning Performance on OOD Prompts for Models Trained with Regression Tasks.} We compare the in-context learning performance of the vanilla-trained model (labeled as \emph{vanilla}) and models trained with auxiliary loss to form task vectors at different layers $l$ (labeled as \emph{ours ($l$)}). We evaluate the models on OOD prompts, including: 
    (a) logistic regression, 
    (b) oulier in context,
    (c) noisy input labels with a noise level of 0.3, 
    (d) scaled weights ($\times 3$).}
\end{figure}

\subsection{Indentification of task vector in GINC dataset}
\begin{figure}[h!]
    \centering
    \includegraphics[width=0.8\linewidth]{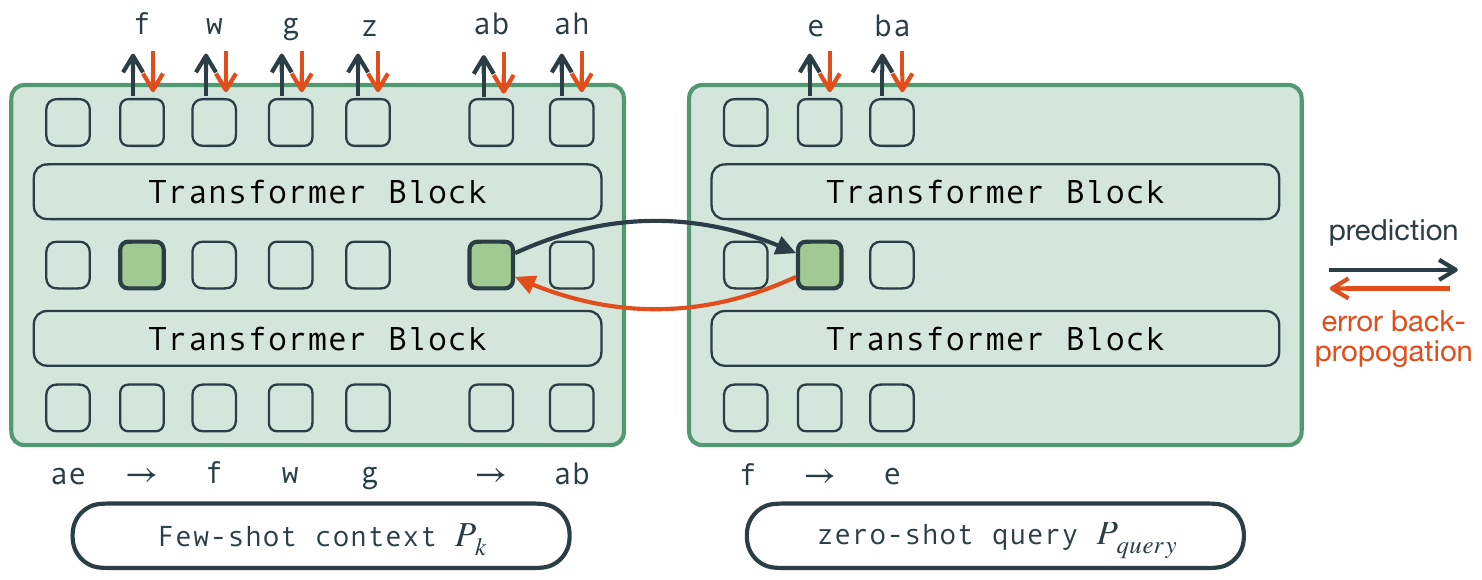}
    \caption{\small \emph{Demonstration of our training algorithm in GINC dataset.} In contrast to the Meta-ICL format described in Figure~\ref{fig:method}, in GINC dataset, there are no explicit input-output pairs. Instead, the entire trajectory performs one single task defined by a hidden markov model. Therefore, we randomly insert several $\rightarrow$'' tokens into the trajectory in the few-shot mode, while inserting one $\rightarrow$'' token into the trajectory in the zero-shot mode. The task vector at the $l$-th layer is then injected into the embedding in the zero-shot mode.}
    \label{fig:ginc_training}
\end{figure}

In GINC, there are no explicit input-output pairs; instead, the entire trajectory performs one task (hidden markov model). Therefore, we encode the task vector into the inserted token ``\texttt{-}'' in the trajectory. We provide an illustration of our training algorithm in Figure.~\ref{fig:ginc_training}.

\end{document}